\documentclass{article} % For LaTeX2e
\usepackage{iclr2024_conference,times}

\usepackage{amsmath,amsfonts,bm}

\def\eqref#1{equation~\ref{#1}}
\def\1{\bm{1}}

\DeclareMathAlphabet{\mathsfit}{\encodingdefault}{\sfdefault}{m}{sl}
\SetMathAlphabet{\mathsfit}{bold}{\encodingdefault}{\sfdefault}{bx}{n}

\usepackage{hyperref}
\usepackage{url}
\usepackage{xspace}
\usepackage{graphicx}
\usepackage{caption}
\usepackage{multirow}
\usepackage{amsmath}
\usepackage{amssymb}
\usepackage{booktabs}
\usepackage{times}
\usepackage{epsfig}
\usepackage{enumitem}
\usepackage[misc]{ifsym}
\usepackage{makecell}
\usepackage{cases}
\usepackage{xcolor}
\usepackage{wrapfig}
\usepackage{multicol}
\usepackage{tikz}

\newcommand{\nickname}{EgoLM}

\title{\nickname: Multi-Modal Language Model of Egocentric Motions}

\author{
Fangzhou Hong\textsuperscript{\rm 1,2},\quad
Vladimir Guzov\textsuperscript{\rm 1,3},\quad
Hyo Jin Kim\textsuperscript{\rm 1},\quad
Yuting Ye\textsuperscript{\rm 1}\\
\textbf{
Richard Newcombe\textsuperscript{\rm 1},\quad
Ziwei Liu\textsuperscript{\rm 2~\Letter},\quad
Lingni Ma\textsuperscript{\rm 1~\Letter}}\\
\textsuperscript{1}Meta Reality Labs Research, \quad
\textsuperscript{2}S-Lab, Nanyang Technological University\\
\textsuperscript{3}University of Tuebingen
}

\newcommand{\ie}{\emph{i.e.}\xspace}
\newcommand{\eg}{\emph{e.g.}\xspace}

\iclrfinalcopy % Uncomment for camera-ready version, but NOT for submission.
\begin{document}

\onecolumn{%
\renewcommand\twocolumn[1][]{#1}%
\maketitle
\begin{center}
    \centering
    \captionsetup{type=figure}
    \includegraphics[width=\textwidth]{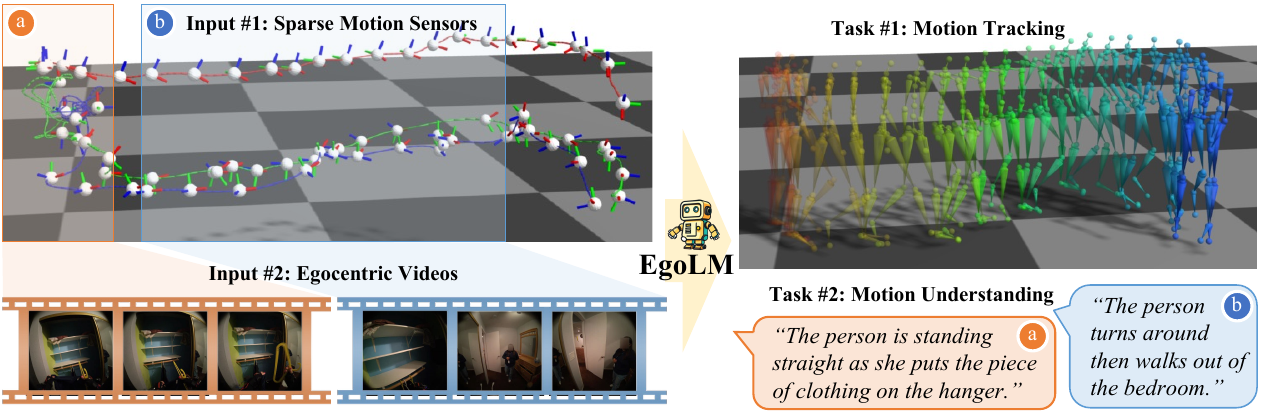}
    \captionof{figure}{We propose \textbf{\nickname{}}, a multi-modal language model that unifies egocentric motion tracking and understanding from wearable sensor data, \eg, sparse motion sensors and egocentric videos.}
    \label{fig:teaser}
\end{center}
}

\begin{abstract}
  % Learning human motions from wearable devices is essential for contextual AI to perform as an everyday smart assistant. To this end, we present \nickname, which unifies tasks of human motion tracking and understanding from egocentric multi-modal inputs, \ie, egocentric videos, motion sensors. To facilitate efficient multitask learning, we build from language models to model the joint distribution of human motion and language. Then, sensor inputs are encoded and projected to the joint latent space and used to prompt motion generation or text generation for motion tracking or understanding. We perform experiments on a large-scale multi-modal human motion dataset to validate the effectiveness of our proposed multitask multi-modal learning.

  As the prevalence of wearable devices, learning egocentric motions becomes essential to develop contextual AI. In this work, we present \textbf{\nickname}, a versatile framework that \textbf{tracks} and \textbf{understands} egocentric motions from \textbf{multi-modal} inputs, \eg, egocentric videos and motion sensors.
  \nickname{} exploits rich contexts for the disambiguation of egomotion tracking and understanding, which are ill-posed under single modality conditions.
  To facilitate the versatile and multi-modal framework,
  our key insight is to model the joint distribution of egocentric motions and natural languages using large language models (LLM).
  % with newly proposed \textit{motion tokenization}, \textit{motion pre-training} and \textit{multi-modal instruction tuning}.
  Multi-modal sensor inputs are encoded and projected to the joint latent space of language models, and used to prompt motion generation or text generation for egomotion tracking or understanding, respectively.
  Extensive experiments on large-scale multi-modal human motion dataset validate the effectiveness of \nickname{} as a generalist model for universal egocentric learning. Project Page: \url{https://hongfz16.github.io/projects/EgoLM}.
\end{abstract}

\section{Introduction} \label{sec:intro}

With the recent explosive advancement of large language models, their values as intelligent agents have been thoroughly studied~\citep{gpt2,gpt3,gpt4,llama,llama2}. To better play the role of everyday smart assistant, the contextualization of AI is proposed and studied~\citep{vercauteren2019cai4cai,deepika2020jollity}. Agents are expected to interact with users in a context-aware style, through multi-modal sensors on wearable devices, \eg, smartwatches, smart glasses~\citep{projectaria}.
Human motions play an important role in the user-agent interaction~\citep{outlook}, which requires egocentric human motion learning~\citep{li2015delving}.

In this work, we propose a versatile framework \textbf{\nickname} that approaches human motions learning from egocentric perspective.
Specifically, \nickname{} unifies two aspects of egocentric motion learning, \ie, tracking and understanding.
\textbf{a)} Egocentric motion tracking aims to recover full-body motions from sparse motion sensors, \eg, three-points (head and both wrists) 6-DoF poses~\citep{avatarposer,bodiffusion,avatargrowlegs,jiang2023egoposer} or one-point (only head) 6-DoF poses~\citep{egoego}.
\textbf{b)} Egocentric motion understanding aims to recognize or describe human motions from wearable sensors, \eg, egocentric videos~\citep{Damen2021PAMI,Damen2022RESCALING,Damen2018EPICKITCHENS,nagarajan2024egoenv,xue2023egocentric,escobar2022video,grauman2022ego4d,rodin2021predicting,yonetani2016recognizing,del2016summarization,chen2023egoplan}.
% In our setting, we aim at step-by-step narration for human motions.
Both tasks are highly challenging due to the \textbf{incomplete observation from egocentric perspectives}. To this end, we propose to approach the challenges in an unified way by incorporating \textbf{multiple modalities} and \textbf{multi-task training}, which are elaborated below.

Egocentric motion tracking from sparse sensors is an ill-posed problem. Three-points~\citep{avatarposer} and one-point inputs~\citep{egoego} miss information of the lower body parts and even hand positions, making it a one-to-many mapping problem.
In order to disambiguate the tracking of unobserved body parts, we explore the environment contexts by egocentric videos captured from head-mounted cameras.
Although other body parts are not always visible from egocentric videos, the semantics of the environment provides valuable clues to disambiguate full body motions.

For egocentric motion understanding, the common input setting is the egocentric video~\citep{jia2022egotaskqa}. However, egocentric videos lacks the accurate information of full-body motion, for their restricted viewing angles. For better understanding of human motion, sparse motion sensor data is valuable in terms of providing accurate body part positions~\citep{tan2024egodistill}.
Therefore, we unify the input conditions as egocentric videos and sparse sensors for both tracking and understanding. Further unifying the training of both tasks can also be beneficial, especially for the understanding part. The supervision signals of full-body motions from motion tracking training can contribute to motion understanding.

% Both tasks require egocentric video and motion sensor data as input. Unification of both tasks can also benefit each other.
% => in summary, we aim at a multi-modal multi-tasking framework. To facilitate this, our key insight is to use language models for their versatility and scalability. Two key challenges: large modality gaps, large task gaps.

In summary, we aim at a multi-modal multi-tasking generative framework. As shown in Fig.~\ref{fig:teaser}, \nickname{} takes sparse motion sensor data (three-points or one-point) and egocentric videos as inputs. Then motion and natural languages are generated for motion tracking and understanding, respectively. To facilitate this versatile framework, there are two main challenges in the framework design, which are \textbf{large modality gaps} and \textbf{large task gaps}. To that end, our key insight is to \textbf{use a language model to handle the multi-modal inputs and multi-task training}.

% Recent trends in multi-modal learning, \ie, Vision Language Models~\citep{liu2023llava,liu2023improvedllava}, shows the scalability and versatility of language models. They typically work with two modalities, \ie images and texts.
Unlike recent VLMs~\citep{liu2023llava,liu2023improvedllava}, our setting is more complex and challenging, where four modalities are involved, including sparse motion sensor data, egocentric videos, motion representations, and texts. These modalities provide different granularity of information. Human motions and sparse motion sensor data are low-level and contiguous representations with physical meanings. Natural languages, on the other hand, are unstructured and discrete representations. To bridge the gap, we adopt three strategies:
\textbf{a)} Treat motions as languages. A motion VQ-VAE is trained to tokenize motions, which can be generated autoregressively by a language model.
\textbf{b)} Unify different inputs to the language model space. Sparse sensor data and egocentric videos are encoded and projected by light-weight temporal encoders.
\textbf{c)} Use instruction tuning for multi-task joint training.

% The first challenge is the \textbf{large gap between motion representations and natural languages}, which might leads to difficulties in multitask learning. Human motions are low-level and contiguous representations with physical meanings. Natural languages, on the other hand, are unstructured and discrete representations. To bridge the gap, we take inspiration from recent works of joint motion-language modeling~\citep{jiang2024motiongpt,zhang2023motiongpt}, where motions are treated as a foreign language and modeled by language models. Specifically, we first learn a motion VQ-VAE as the tokenizer to discretize the contiguous motion representations to tokens. Then, we perform motion pre-training with a language model. The model would capture the distributions of both motion and natural languages, which addresses the challenge of unifying motion and natural languages.

% The second challenge is the \textbf{multitask multi-modal learning}. Our input modalities include egocentric videos and motion sensor data, which are very different in terms of data size and information intensity. Therefore, to unify these modalities and tasks, we take inspiration from recent advancements in vision language models~\citep{liu2023llava,liu2023improvedllava}. Different modalities are encoded separately and projected to the language model embedding space. The training of different tasks are unified by instruction tuning, where tasks are distinguished by instructions.

To validate the proposed framework, we perform extensive experiments on a large-scale motion dataset, Nymeria~\citep{ma2024nymeria}. Compared with previous motion tracking and understanding methods, our newly proposed multi-modal setup shows its advantages. Our contributions are summarized below.

\noindent\textbf{1)} We propose a versatile multi-modal generative framework, \nickname{}, that unifies egocentric motion tracking and understanding tasks with a language model.

\noindent\textbf{2)} A new egocentric motion tracking setup is proposed. We combine sparse sensor inputs with egocentric videos to provide more contexts that disambiguate this ill-posed problems.

\noindent\textbf{3)} We propose a practical paradigm of motion understanding by combining sparse sensor data and egocentric videos, which provides more accurate full-body motion narration.

\noindent\textbf{4)} Extensive experiments and studies are performed to show the effectiveness of the proposed framework. Our setup achieves the best performance compared with the state-of-the-art methods.

\section{Related Work} \label{sec:related}

% \subsection{Human Motion Learning}

\noindent\textbf{Motion Regression.}
Large amounts of efforts are devoted to detect and track 2D or 3D keypoints from human images and videos~\citep{toshev2014deeppose,martinez_2017_3dbaseline,pavllo:videopose3d:2019}. To incorporate more human structure prior, parametric human models, \eg, SMPL~\citep{loper2023smpl}, are used as the regression target~\citep{Bogo:ECCV:2016,hmrKanazawa17}. Other than the cameras, wearable motion sensors are straight-forward in terms of motion capture~\citep{ponton2023sparseposer,mollyn2023imuposer,milef2023variational,yi2023egolocate,jiang2023egoposer}. Recent advancements in VR/AR devices and applications have developed a new setup for motion tracking, \ie, three-points body tracking~\citep{avatargrowlegs,avatarposer,bodiffusion}. EgoEgo~\citep{egoego} proposes to track motions from only head poses. In this work, we also target motion tracking from sparse sensors. The difference is that we propose to use egocentric videos to disambiguate ill-posed scenarios in this setup.

\noindent\textbf{Motion Generation.}
There have been many efforts in generating motions from various conditions, \ie, action labels~\citep{petrovich2021action,guo2020action2motion}, natural languages~\citep{zhang2024motiondiffuse,tevet2022human,BABEL:CVPR:2021,Guo_2022_CVPR,zhang2023t2m,guo2022tm2t}. Recently, researchers take advantage of powerful LLMs to model the joint motion-language distribution for text-to-motion generation~\citep{jiang2024motiongpt,zhang2023motiongpt,zhou2023avatargpt}. In this work, we also adopt the similar idea of modeling motion together with language models. As a by-product, we can also perform text-to-motion generation. But our main focus is on motion tracking and understanding from multi-modal inputs.

\noindent\textbf{Motion Understanding.}
There have been many different setups in motion understanding. From the input side, human videos, either from third-person view~\citep{soomro2012ucf101,kuehne2011hmdb,Tran_2015_ICCV,wang2016temporal,yan2018spatial} or first-person view~\citep{Damen2021PAMI,Damen2022RESCALING,Damen2018EPICKITCHENS}, are used for this task. From the output side, action recognition/classification has been a classic task definition~\citep{soomro2012ucf101,Damen2018EPICKITCHENS}. More recently, with the development of language models, some researches also propose to use natural languages as output~\citep{jia2022egotaskqa,xu2024retrieval,grauman2022ego4d,xue2023egocentric,chen2023egoplan}. In our work, from the input side, we propose to combine egocentric videos, which provide high-level semantic information, with motion sensor inputs, which carries low-level motion clues, for more holistic motion understanding. For the output, we use natural language responses for more versatility and diversity.

% \subsection{Language Models}

\noindent\textbf{Language Models.}
Language models have been a huge success in recent years with the large-scale pre-training~\citep{gpt2,gpt3} and alignment~\citep{chatgpt,gpt4}. To take advantage of the powerful text generation ability, image~\citep{liu2023llava,liu2023improvedllava} or video understanding~\citep{zhang2023video} are defined as conditional text generation. LLaVA~\citep{liu2023llava} proposes to encode images with powerful pre-trained vision encoders~\citep{radford2021learning} and inject the features to language models. By tuning from a powerful LLM~\citep{llama}, LLaVA achieves wonderful abilities of vision question answering. In this work, we also adopt the similar idea of encoding and injecting features of other modalities in the language model and unifying different tasks with instruction tuning.

\section{Method} \label{sec:methods}

The overview of \nickname{} is demonstrated in Fig.~\ref{fig:overview}. There are three steps in \nickname{} training. In the first step, we train a motion VQ-VAE as the motion tokenizer (Sec.~\ref{sec:motion_tok}). The second step is motion pre-training for motion distribution learning (Sec.~\ref{sec:motion_pre}). The last step is multi-modal instruction tuning to guide the model to perform motion tracking and understanding (Sec.~\ref{sec:motion_ins}).
% and will be elaborated in this section. Specifically, we first briefly describe essential techniques of language models and motion representations in Sec.~\ref{sec:preliminaries}. To utilize language models for motion modeling, we design a motion tokenizer by training an expressive motion VQ-VAE, which is discussed in Sec.~\ref{sec:motion_tok}. Then we use next-token prediction to pre-train the language model to learn the motion distribution in Sec.~\ref{sec:motion_pre}. Lastly, we perform multi-modal instruction tuning for egocentric motion tracking and understanding in Sec.~\ref{sec:motion_ins}.

\subsection{Preliminaries} \label{sec:preliminaries}

\noindent\textbf{Language Model.}
Language models model the distribution of natural languages. Recent breakthroughs in language models suggest the effectiveness of the transformer-based architecture~\citep{vaswani2017attention}. The language model consists of three parts. The first is a look-up table (LM embedding) that stores the embeddings for each text token. The second part is the transformer backbone that takes text embeddings as inputs. The output features are mapped to probabilities of the next tokens by the third part of LM head.

\noindent\textbf{Motion Representation.}
Human motions are represented as sequences of poses, global translations and rotations defined on the root joint. Each frame of pose is represented by joint angles, defined on a kinematic tree. For better learning of motion dynamics, we also include joint angle velocity in the representation. To avoid the normalization of global translation, we use the translation velocity $V_t^r\in\mathbb{R}^3$ for each frame, which can be integrated back to global translations. To ease the regression difficulty of rotation angles, we use 6D rotation representations~\citep{hempel20226d} for the root rotation $R_t^r\in\mathbb{R}^6$, root rotation velocity $R_t^{rv}\in\mathbb{R}^6$, joint angles $R_t^j\in\mathbb{R}^{22\times6}$, and joint angle velocity $R_t^{jv}\in\mathbb{R}^{22\times6}$. Formally, we represent human motions with $T$ frames as $M = \{P_t\}_{t=1}^{T}$, where $P_t = [V_t^r; R_t^r; R_t^{rv}; R_t^j; R_t^{jv}] \in \mathbb{R}^{279}$.
% \begin{equation}
%     P_t = [V_t^r; R_t^r; R_t^{rv}; R_t^j; R_t^{jv}].
% \end{equation}
Forward kinematics (FK) together with integration of root velocity can be used to recover the joint positions $J = \text{FK}(M)\in\mathbb{R}^{23\times3}$.
% TODO: can add formal definition for the fk process

\begin{figure}[tb]
  \centering
  \vspace{-15pt}
  \includegraphics[width=\textwidth]{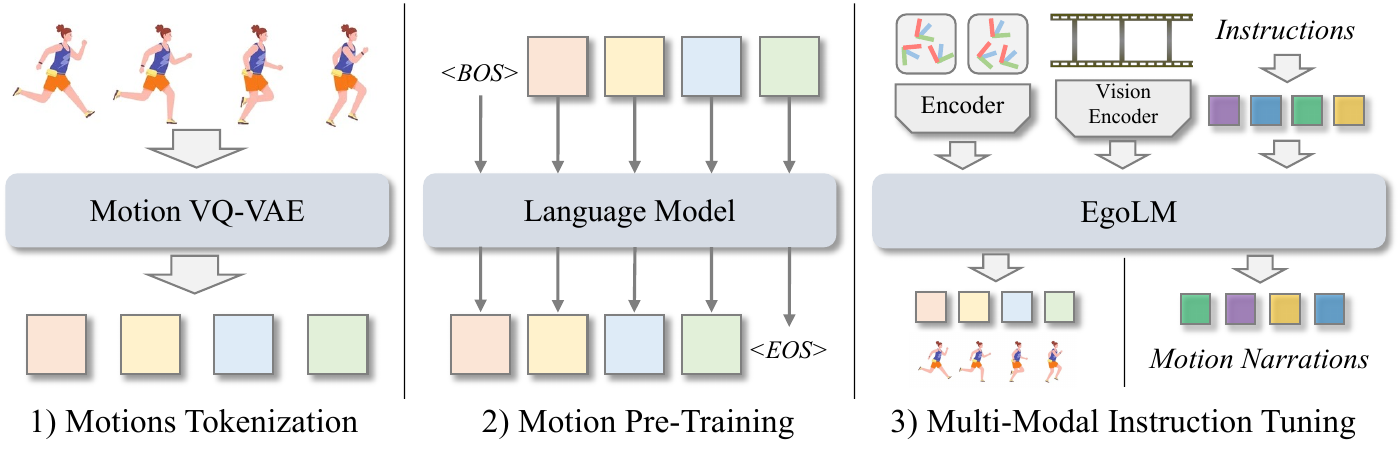}
  \vspace{-20pt}
  \caption{\textbf{Overview of \nickname{}.} Three steps are designed for the training of \nickname{}, \ie, motion tokenizer training, motion pre-training and multi-modal instruction tuning.}
  \vspace{-10pt}
  \label{fig:overview}
\end{figure}

\subsection{Motion Tokenizer} \label{sec:motion_tok}

To treat the motion as a foreign language and train with language models, we first need a motion tokenizer, which can be realized by VQ-VAE~\citep{oord2017neural}.
% To unify the task of motion tracking and understanding, we propose to treat motion a foreign language and learn a joint motion-language distribution using language models. Therefore, similar to text tokenizer, a motion tokenizer is in need. However, unlike natural languages, motions are continuous representation. To overcome it, we train a Vector-Quantized Variational AutoEncoder (VQ-VAE). The VQ-VAE encode maps raw motion representations to discrete tokens, and the decoder of which can recover motion representation from motion tokens. % To ensure high quality motion generation, it is important to design a high-performance VQ-VAE in terms of the reconstruction quality.
%
The motion VQ-VAE consists of a fully convolutional encoder $\mathcal{E}$ and decoder $\mathcal{D}$
% , which are stacks of 1-D convolutions with residual blocks
. The fully convolutional design enables processing motions with arbitrary lengths. The encoder embeds raw motion representation to latent features $f^m = \mathcal{E}(M)$, where $f^m\in\mathbb{R}^{T/r \times c}$, $M\in\mathbb{R}^{T\times279}$. $r$ is the down-sample rate.

Then, codebooks are learned to quantize the motion latent features. We use three techniques in the quantization process, which are 1) exponential moving average (EMA), 2) codebook reset~\citep{dhariwal2020jukebox}, 3) product quantization~\citep{jegou2010product,lucas2022posegpt}. The first two techniques increase the usage rate of codebooks. Product quantization increases the codebook expressiveness by decomposing the latent space into a Cartesian product of sub-spaces with lower dimensions. Specifically, the latent feature $f^m$ is split equally into $N$ trucks $\{f^m_n\}_{n=1}^{N}$, which are quantized separately by $N$ codebooks $\{Z_n\}_{n=1}^{N}$. Each codebook with $K$ entries is defined as $Z_n = \{z_i\}_{i=1}^{K}$, where $z_i\in\mathbb{R}^{c/N}$. The quantization process for feature $f^m_{tn}$ at frame $t$ and trunk $n$ is formulated as
\begin{equation}
     i_{tn} = Q(f^m_{tn}) = \arg\min_{z_i\in Z_n} \|f^m_{tn} - z_i\|_2.
\end{equation}
The resulting indices $i_{tn}$ are flattened and used as motion token sequences $W = \{[(i_n)_{n=1}^{N}]_t\}_{t=1}^{T/r}$, which has the length of $L_W = N\times(T/r)$.
After quantization, we obtain the corresponding codebook entry for the motion latent feature $\hat{f}^m = \{\hat{f}^m_t\}_{t=1}^{T/r} = \{z_{i_t}\}_{t=1}^{T/r}$. It is input into the decoder $\mathcal{D}$ to decode raw motion representation $\hat{M} = \mathcal{D}(\hat{f}^m)$.

For the training of VQ-VAE, two types of training losses are used. The first is the commitment loss $\mathcal{L}_c = \|f^m - \hat{f}^m\|_2$ for the codebook learning. The second is motion reconstruction loss $\mathcal{L}_r$, which consists of raw representation loss $\mathcal{L}_m$, joint position loss $\mathcal{L}_j$, rotation velocity loss $\mathcal{L}_v$, which are defined as
\begin{align}
    \mathcal{L}_r
    &= \lambda_m\mathcal{L}_m + \lambda_j\mathcal{L}_j + \lambda_v\mathcal{L}_v = \lambda_m\|M - \hat{M}\|_1 + \lambda_j\|\text{FK}(M) - \text{FK}(\hat{M})\|_1 \\
    &+ \lambda_v\|R_{1:T-1}^{rv} - (R_{1:T-1}^{r})^{-1}R_{2:T}^{r}\|_1 + \lambda_v\|R_{1:T-1}^{jv} - (R_{1:T-1}^{j})^{-1}R_{2:T}^{j}\|_1.
\end{align}
We define the smoothed L1 loss as $\|\cdot\|_1$. In summary, the training loss of the motion VQ-VAE is $\mathcal{L}_{vq} = \lambda_c\mathcal{L}_c + \lambda_r\mathcal{L}_r$, where $\lambda_*$ are manually adjusted weights.

\subsection{Motion Pre-training} \label{sec:motion_pre}

As discussed before, we build the motion learning framework on a pre-trained language model. However, the pre-trained language models only model the distribution of natural languages. Therefore, to empower them to generate motions, we perform motion pre-training to learn motion distributions. The motion pre-training is conducted similarly to language model pre-training.

Before we can start training the language model, two modifications to the model are needed. Firstly, since the pre-trained language model only contains embeddings for text tokens, we expand the embeddings in accordance with the motion codebook size. Secondly, the output shape of the language model head is also expanded for the same reason. The language model is ready for motion pre-training after the above preparations. Using the motion tokenizer described above, motion representations $M$ can be encoded into a sequence of motion tokens $W = \{w_i\}_{i=1}^{L_W}$. They are fed into the language model to learn the motion token distribution by conducting the classic next-token prediction~\citep{gpt2}. The loss function of this stage $\mathcal{L}_{pre}$ is formulated as
\begin{equation}
    \mathcal{L}_{pre} = - \sum_{i=2}^{L_W}\mathbb{P}(w_i|w_1...w_{i-1}; \Theta), \label{eq:next-token-pred}
\end{equation}
where we maximize the log-likelihood of the next-token probability given the previous token inputs and network parameter $\Theta$.

After the training of this stage, as a by-product, we obtain an unconditional motion generator. Given a leading motion sequence as the prompt, it can autoregressively sample an arbitrary length of reasonable human motion that continues the given motion. More importantly, the language model learns the distribution of human motions and has the ability of sampling plausible human motions, which lays a solid foundation for the next stage.

\begin{figure}[tb]
  \centering
  \vspace{-15pt}
  \includegraphics[width=\textwidth]{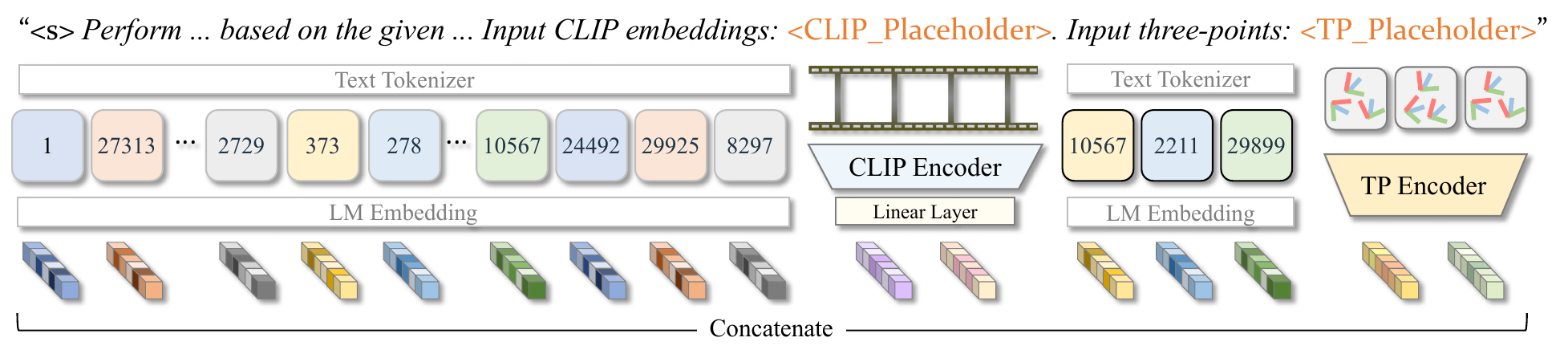}
  \vspace{-20pt}
  \caption{\textbf{Details of Multi-Modal Instruction Tuning.} Different modalities are encoded separately. Their features are concatenated in the order of the instruction template and input into the transformer layers of the language model.}
  \label{fig:stage3_detail}
  \vspace{-5pt}
\end{figure}

\subsection{Multi-Modal Instruction Tuning} \label{sec:motion_ins}

Inspired by recent advancements in LLMs~\citep{gpt4,chatgpt,zheng2023judging}, to squeeze the power out of generative pre-training models, instruction tuning is adopted to guide models with instructions to perform specific tasks. The instruction template usually consists of 1) instructions that specify which tasks to perform; 2) inputs of the task; 3) outputs. We also envision our model accepting multi-modal sensor data as inputs. However, even with motion pre-training, the model only accepts text or motion tokens as inputs. It is not practical or necessary to design tokenizers and perform pre-training for all the involved modalities. Therefore, we draw inspiration from vision language models~\citep{liu2023llava,liu2023improvedllava}, where they directly map vision data to LLM feature space to enable visual question answering.

% We mainly deal with sequential input sensor data, defined as $S\in\mathbb{R}^{T\times C_s}$. They are first encoded by a fully convolutional network to latent features $f_S\in\mathbb{R}^{T/r_S\times C_t}$, where $r_S$ is the down-sample rate, $C_t$ is the transformer input channel size. Then $f_S$ is concatenated with text embeddings of the rest of the instructions and inputs and directly input into the second part of the language model, a transformer-based network.

Specifically, we consider two input modalities other than motion and natural languages, which are egocentric videos and motion sensor inputs. Motion sensor inputs can be three-points (head and wrists) 6-DoF poses or one-point (only head) 6-DoF poses. Both are encoded with positions, velocity, rotation and angular velocity. Below, we use three-points as examples. We unify both motion tracking and motion understanding using the following templates.

% \vspace{5pt}
\noindent\begin{tabular}{cc}
\centering
\footnotesize
\fbox{
\parbox{.43\textwidth}{
\textbf{Task:} \textit{Motion Tracking} \\
\textbf{Instruction:} \textit{Perform motion tracking based on the given three-points and CLIP embeddings.} \\
\textbf{Input:} \textit{Input CLIP embeddings:} \texttt{<CLIP\_Placeholder>}. \textit{Input three-points feature:} \texttt{<TP\_Placeholder>} \\
\textbf{Output:} \texttt{<Motion\_Placeholder>}
}}
&
\centering
\footnotesize
\fbox{
\parbox{.46\textwidth}{
\textbf{Task:} \textit{Motion Understanding} \\
\textbf{Instruction:} \textit{Describe the human motion based on the given three-points and CLIP embeddings.} \\
\textbf{Input:} \textit{Input CLIP embeddings:} \texttt{<CLIP\_Placeholder>}. \textit{Input three-points feature:} \texttt{<TP\_Placeholder>} \\
\textbf{Output:} \texttt{<Narration\_Placeholder>}
}}
\end{tabular}

% \vspace{5pt}
% {\centering
% \footnotesize
% \quad
% \fbox{
% \parbox{110mm}{
% \textbf{Task:} Motion Tracking \\
% \textbf{Instruction:} Perform motion tracking based on the given three-points and CLIP embeddings. \\
% \textbf{Input:} Input CLIP embeddings: \texttt{<CLIP\_Placeholder>}. Input three-points feature: \texttt{<TP\_Placeholder>} \\
% \textbf{Output:} \texttt{<Motion\_Placeholder>}
% }}}

% \vspace{5pt}
% {\centering
% \footnotesize
% \quad
% \fbox{
% \parbox{110mm}{
% \textbf{Task:} Motion Understanding \\
% \textbf{Instruction:} Describe the human motion based on the given three-points and CLIP embeddings. \\
% \textbf{Input:} Input CLIP embeddings: \texttt{<CLIP\_Placeholder>}. Input three-points feature: \texttt{<TP\_Placeholder>} \\
% \textbf{Output:} \texttt{<Narration\_Placeholder>}
% }}}
% \vspace{5pt}
% \vspace{5pt}

The encoded three-points 6-DoF poses would replace \texttt{<TP\_Placeholder>}. \texttt{<CLIP\_Placeholder>} is the placeholder for egocentric video features. Motion tokens are filled in \texttt{<Motion\_Placeholder>}. \texttt{<Narration\_Placeholder>} is the placeholder for corresponding motion narration. A detailed illustration of how we organize different modalities of data is shown in Fig.~\ref{fig:stage3_detail}. Texts are tokenized and translated to feature vectors through LM embedding. Egocentric videos are first encoded by CLIP image encoder~\citep{radford2021learning} per frame, which are further projected by linear layers to the language model feature space. Similarly, motion sensor data, \eg, sequences of three-points 6-DoF poses, is encoded by a fully convolutional encoder. Lastly, all the encoded features are concatenated and input into the transformer layers of the language model.

For the training of motion understanding, to better learn the joint distribution of motion and natural languages, we also include two auxiliary tasks in the joint instruction training, which are motion-to-text and text-to-motion generation. They are also defined with the templates similar to the above ones. In summary, we train the four tasks jointly as the last step. The loss function is the same next-token prediction loss, as defined in Eq.~\ref{eq:next-token-pred}.

\section{Experiments}
\label{sec:experiments}

\subsection{Experimental Setup}

\noindent\textbf{Dataset.}
We use the Nymeria dataset~\cite{ma2024nymeria} to train and validate our method. The dataset provides \textbf{a)} full body motions, captured by the Xsens Mocap system~\citep{xsens09}, \textbf{b)} egocentric videos, captured by Aria glasses~\citep{projectaria}, and \textbf{c)} narrations of motions written by human annotators. Three-points 6-DoF poses are taken from ground truth joints. For motion tracking, the training set consists of $147.89$h of data and the test set has $41.93$h of data. For motion understanding, the training set has $16673$ segments, each lasting for $3$-$5$ seconds, adding up to $15.77$h. The test set consists of $7468$ segments, $6.76$h of data.

\noindent\textbf{Training Details.}
Motion VQ-VAE has two codebooks, each having $8192$ entries and code dimension of $64$. The down-sample rate is $r=4$. For motion tracking, all experiments are conducted with window size of $60$ frames, which is $1$ second. Random rotation augmentation is applied on motions. We choose to use GPT2-Medium~\citep{gpt2} as the language backbone.

\noindent\textbf{Evaluation Protocols.}
For motion tracking, we calculate joint position errors (for full, upper and lower body), joint angle errors (for full body and root joint).
% We include AvatarPoser~\cite{avatarposer}, BoDiffusion~\cite{bodiffusion} and EgoEgo~\cite{egoego} as baseline methods.
For motion understanding, the outputs are natural languages. Therefore, we adopt NLP metrics, including BERT~\citep{zhang2019bertscore}, BLEU~\citep{papineni2002bleu}, and ROUGE~\citep{lin2004rouge} scores.
% TM2T~\cite{guo2022tm2t} and MotionGPT~\cite{jiang2024motiongpt} are used as baseline methods for motion understanding.

\subsection{Motion Tracking}

\setlength{\tabcolsep}{4.4pt}
\begin{table}[t]
\vspace{-15pt}
  \caption{
    \textbf{Quantitative Results of Motion Tracking.} ``Full'', ``Upper'', ``Lower'' are joint position errors in $mm$. ``J.A.'', ``Root'' are joint angle errors for full body and root joint in degree. $^\dagger$We directly replace three-points with one-point to train AvatarPoser.
  }
  \vspace{-5pt}
  \label{tab:body_tracking}
  \centering
  % \scriptsize
  \begin{tabular}{l|ccc|cccccc}
    \toprule
    \multirow{2}{*}{Method} & \multicolumn{3}{c|}{Input Modality} & \multirow{2}{*}{Full} & \multirow{2}{*}{Upper} & \multirow{2}{*}{Lower} & \multirow{2}{*}{J.A.} & \multirow{2}{*}{Root} \\
    % \cline{2-4}
    & 3pts & 1pt & Vid. & & & & & & \\
    \midrule
    AvatarPoser~\citep{avatarposer} & \checkmark & & & 85.89 & 52.78 & 165.18 & \textbf{12.41} & 14.78 \\
    Bodiffusion~\citep{bodiffusion} & \checkmark & & &  79.80 & 52.79 & 152.68 & 12.74 & \textbf{13.09} \\
    Ours & \checkmark & & & 83.88 & 54.06 & 148.37 & 13.31 & 14.13 \\
    % \hline
    \midrule
    Ours & \checkmark & & \checkmark & \textbf{73.38} & \textbf{49.67} & \textbf{124.58} & 12.48 & 13.23 \\
    \midrule
    AvatarPoser$^\dagger$~\citep{avatarposer} & & \checkmark & & 129.23 & 94.19 & 192.34 & 16.55 & 21.60 \\
    EgoEgo~\citep{egoego} & & \checkmark & & 132.16 & 100.02 & 190.32 & 18.90 & 21.80 \\
    Ours & & \checkmark & & 127.45 & 97.87 & 174.92 & 16.97 & 20.57 \\
    % \hline
    \midrule
    Ours & & \checkmark & \checkmark & \textbf{106.95} & \textbf{83.73} & \textbf{141.26} & \textbf{14.67} & \textbf{19.04} \\
  \bottomrule
  \end{tabular}
  % \vspace{-15pt}
\end{table}

\noindent\textbf{Quantitative Results.}
We report quantitative results of motion tracking in Tab.~\ref{tab:body_tracking}. All methods are evaluated with batch inference, meaning that every $60$ frames are inferenced independently. We evaluate several different input combinations of three modalities, which are three-points 6-DoF poses (``3pts''), one-point 6-DoF poses (``1pt'') and egocentric videos (``Vid''). For the 3pts-only and 1pt-only settings, \nickname{} achieves comparable performance with baseline methods. This show the effectiveness of using language models to perform precise motion tracking tasks.
Moreover, we also use egocentric videos to provide environment contexts for motion tracking. For three-points tracking, the additional modality brings $10mm$ improvement in full body joints error. For the one-point tracking, adding egocentric videos improves joints error by $20mm$. It shows the effectiveness of using egocentric videos as context information for disambiguation of the ill-posed problem.
% Since we do not assume that the floor height is known, we normalize the three-points or one-point inputs by moving all other frames to the coordinate of the first frame. This causes ambiguity in terms of whether the person is standing or seated. egocentric videos help disambiguate such situations, collapsing the sampling space of lower body. For the one-point motion tracking, the input does not contain any information that can infer hand positions, which makes the hand position prediction very unreliable (\textcolor{red}{TODO}hand position error). egocentric videos sometimes capture hands when they are in front of the body, which provides valuable clues for hand position prediction. This explains the large improvement of $47mm$ in terms of hand position error.

\begin{figure}[tb]
  \centering
  \vspace{-5pt}
  \includegraphics[width=\textwidth]{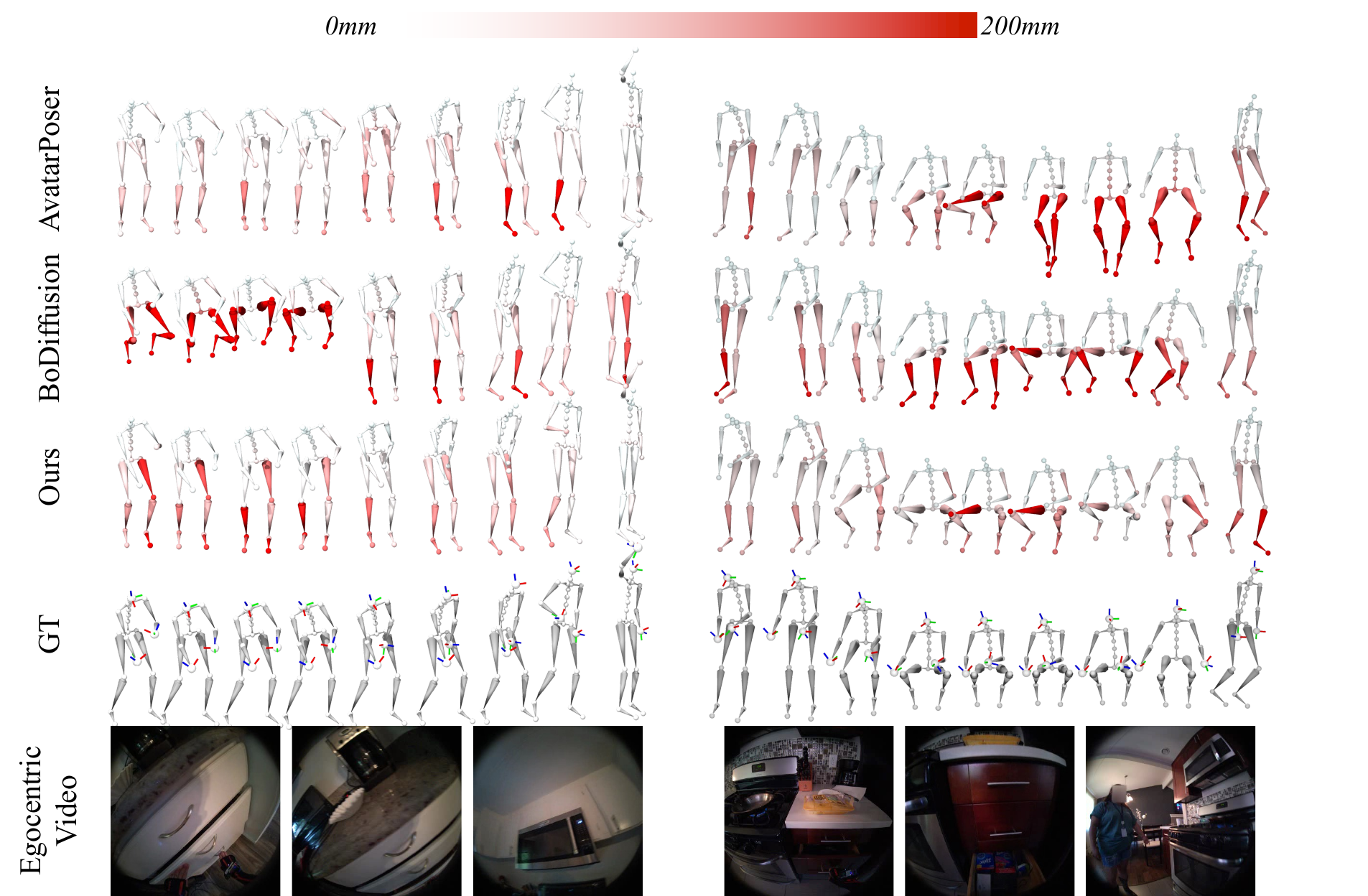}
  \caption{\textbf{Qualitative Results of Three-Points Motion Tracking.} Skeletons are color-coded by the joint position errors. Baseline methods only use three-points as inputs. Ours uses three-points and egocentric videos as inputs.}
  \label{fig:body_tracking}
  \vspace{-15pt}
\end{figure}

\noindent\textbf{Qualitative Results.}
Three-points motion tracking results and comparisons are shown in Fig.~\ref{fig:body_tracking}. Due to the ambiguity of three-points, AvatarPoser mistakenly generates standing poses for squatting sequences (right example). BoDiffusion, for its generative nature, can sample correct results in some cases, \eg, the squatting example. But it also suffers from the ambiguity issue, as shown in the bending down sequence (left example). They show the importance of considering contexts in the motion tracking task for the purpose of disambiguation.
% And we show that CLIP embeddings of egocentric are effective environment contexts.
Our full model can reliably perform three-points body tracking for the shown challenging cases.

\begin{figure}[tb]
\vspace{-20pt}
  \centering
  \includegraphics[width=\textwidth]{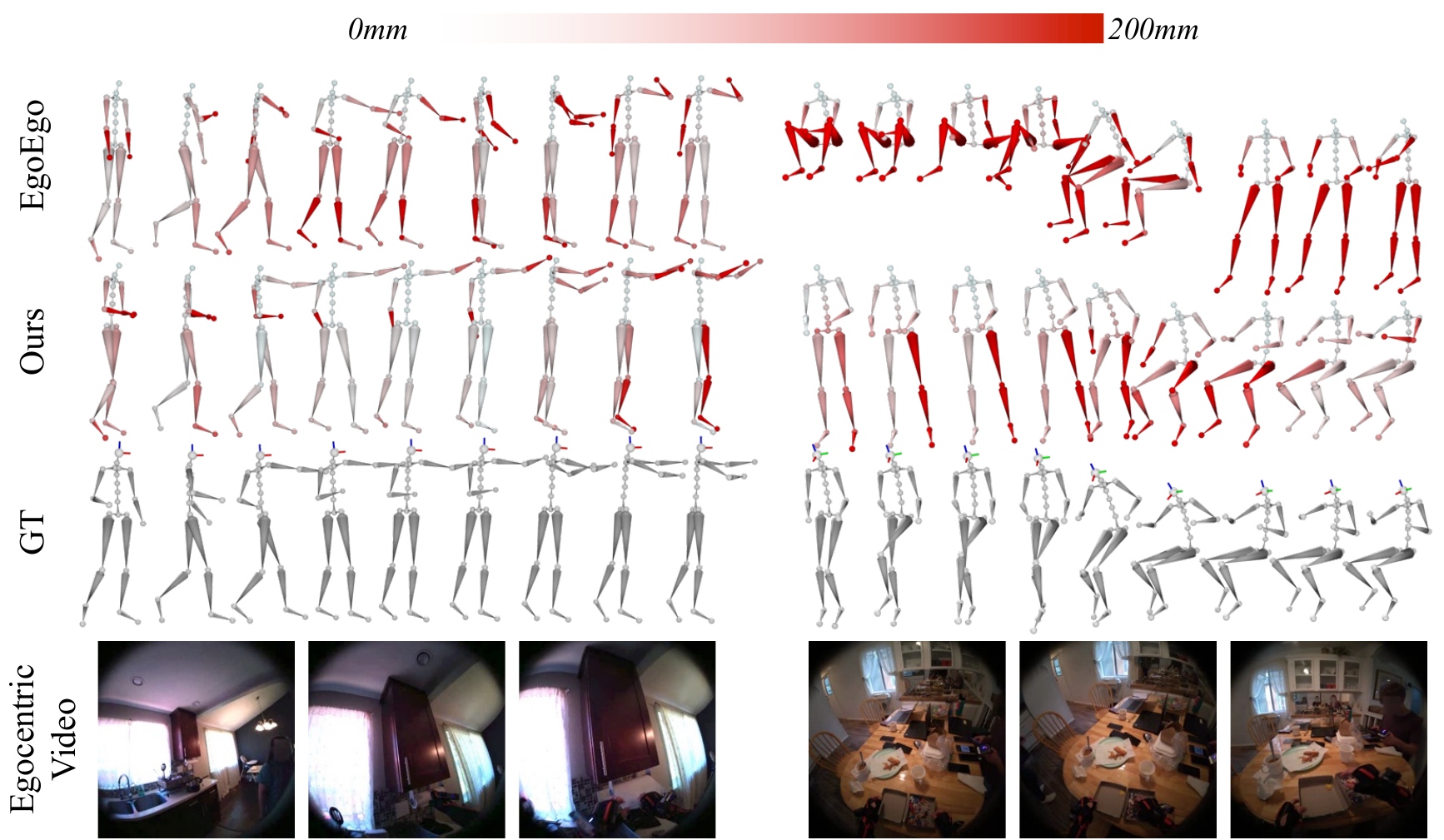}
  \caption{\textbf{Qualitative Results of One-Point Motion Tracking.} Skeletons are color-coded by joint position errors. EgoEgo only uses one-point as inputs. Ours includes egocentric videos as inputs.}
  \label{fig:1pt_body_tracking}
  \vspace{-10pt}
\end{figure}

One-point motion tracking results are shown in Fig.~\ref{fig:1pt_body_tracking}. It is a more challenging task especially for upper body. As shown in the left example, the upper body motions generated by EgoEgo are completely different from the ground truth. In the right example, EgoEgo wrongly generates sitting poses for standing frames and standing poses for sitting frames, which is caused by the ambiguity problem. Egocentric videos in this task not only help to eliminate the ambiguity, but also provide some clues about the hand position. In the left example, when hands are visible in the frames, our model captures this information through CLIP embeddings and generates correct arm movements.

\subsection{Motion Understanding}

\noindent\textbf{Quantitative Results.}
We report quantitative results of motion understanding in Tab.~\ref{tab:motion_understanding}. For this task, we tested three input modalities, \ie, three-points (``3pts''), motions, and egocentric videos (``Vid''). Different combinations of these modalities are evaluated. We first test and compare with two motion understanding methods that only take motion as inputs, TM2T~\citep{guo2022tm2t} and MotionGPT~\citep{jiang2024motiongpt}. TM2T trains language generation from scratch, which explains its poor performance. MotionGPT uses a pre-trained T5 model~\citep{raffel2020exploring}. \nickname (M2T\&T2M) achieves the best performance for the scalability advantage brought by the decoder-only architecture.

% Using motion as inputs requires accurate motion tracking in the first place, which is not always available. How about directly using sensor inputs? To answer this question, we tested the variant of only using three-points (TP2T) and only using egocentric videos (V2T). Compared with the motion-only version, A clear performance drop from TP2T variant can be observed, which is expected since three-points only provide partial information about body motion.
% The performance of egocentric video variant (V2T) is better than the motion-only version. Because our target motion narrations is relevant to environment contexts, which can be inferred from egocentric videos. This also shows the value of using egocentric videos in motion understanding. If further combining egocentric videos with motions as inputs (MV2T\&T2M), which provides all necessary information needed for motion narration and understanding, we achieve the best performance across all variants.

Using motion as inputs requires precise motion tracking, which is not always available. So, we explored using sensor inputs instead. We tested two variants: three-points-only (TP2T) and egocentric videos only (V2T). The TP2T variant showed a noticeable drop in performance compared to the motion-only version, as three-points provide limited information about body motion. In contrast, the V2T variant outperformed the motion-only version because egocentric videos capture environmental context relevant to our motion narrations. This highlights the importance of egocentric videos in understanding motion.
% When we combine egocentric videos with motion inputs (MV2T\&T2M), we achieve the best overall performance, as it provides all the necessary information for motion narration and understanding.

We then test our proposed setup of combing three-points and egocentric videos for motion understanding. There are three ways of achieving this setup. The first one is to combine two existing setups: 1) three-points motion tracking and 2) motion-to-text generation (TPV2M $+$ MV2T). The performance of this variant slightly drops compared with MV2T variant, due to error accumulation.
% We can first use three-points and egocentric videos to predict full body motions, which is then combined with videos for text generation. The performance is not promising, due to error accumulation.
The second way is directly training three-points plus egocentric video to text generation (TPV2T) with the proposed multi-modal instruction tuning. It is better than only using egocentric videos or motions. However, it still falls behind MV2T variant for the missing lower body information. To solve that, we propose to also include three-points motion tracking in training to actively establish the connection between three-points and motion narrations.
% Specifically, for the ``Full'' version, we design four tasks that are trained jointly, three-points $+$ video $\xrightarrow{}$ motion, motion $+$ video $\xrightarrow{}$ narration, narration $\xrightarrow{}$ motion, three-points $+$ video $\xrightarrow{}$ narration. As shown in the last row of Tab.~\ref{tab:motion_understanding},
Joint training improves motion understanding from three-points plus egocentric video, which proves the effectiveness of using motion as a bridge between different modalities.

% \setlength{\tabcolsep}{4pt}
% \renewcommand{\arraystretch}{1.2}
% \begin{table}[t]
%   \caption{
%     \textbf{Motion Understanding Results.}
%   }
%   \label{tab:motion_understanding}
%   \centering
%   \scriptsize
%   \begin{tabular}{c|ccc|ccccc}
%   \toprule
%   \multirow{2}{*}{Method} & \multicolumn{3}{|c|}{Input Modality} & \multirow{2}{*}{Bert} & \multirow{2}{*}{Bleu@1} & \multirow{2}{*}{Bleu@4} & \multirow{2}{*}{CIDEr} & \multirow{2}{*}{RougeL} \\
%   \cline{2-4}
%   & 3pts & Motion & Vid. & & & & & \\
%   \midrule
%   TM2T~\cite{} & & \checkmark & & 11.08 & 40.11 & 8.99 & 20.85 & 30.70 \\
%   MotionGPT~\cite{} & & \checkmark & & 14.09 & 42.22 & 10.31 & 37.27 & 32.33 \\
%   \hline
%   Ours (M2T\&T2M) & & \checkmark & & 15.90 & 42.68 & 11.06 & 25.74 & 33.71 \\
%   Ours (TP2T) & \checkmark & & & 11.94 & 41.70 & 9.85 & 17.11 & 31.47 \\
%   Ours (V2T) & & & \checkmark & 16.62 & 43.03 & 11.34 & 35.53 & 33.13 \\
%   Ours (MV2T\&T2M) & & \checkmark & \checkmark & 20.32 & 45.33 & 12.80 & 45.50 & 35.31 \\
%   \hline
%   Ours (TPV2M, MV2T) & \checkmark & & \checkmark & 19.97 & 45.41 & 12.81 & 41.38 & 35.04 \\
%   Ours (TPV2T) & \checkmark & & \checkmark & 18.38 & 44.55 & 12.12 & 34.54 & 33.80 \\
%   Ours (Full) & \checkmark & & \checkmark & 19.40 & 45.45 & 12.72 & 26.13 & 34.82 \\
%   \bottomrule
%   \end{tabular}
% \end{table}

\setlength{\tabcolsep}{3.2pt}
\begin{table}[t]
\vspace{-20pt}
  \caption{
    \textbf{Quantitative Results of Motion Understanding.} Different input modality combinations are tested. All metrics are higher the better.
  }
  \vspace{-5pt}
  \label{tab:motion_understanding}
  \centering
  % \scriptsize
  \begin{tabular}{l|ccc|cccc}
  \toprule
  \multirow{2}{*}{Method} & \multicolumn{3}{c|}{Input Modality} & \multirow{2}{*}{Bert$\uparrow$} & \multirow{2}{*}{Bleu@1$\uparrow$} & \multirow{2}{*}{Bleu@4$\uparrow$} & \multirow{2}{*}{RougeL$\uparrow$} \\
  % \cline{2-4}
  & 3pts & Motion & Vid. & & & & \\
  \midrule
  TM2T~\citep{guo2022tm2t} & & \checkmark & & 11.08 & 40.11 & 8.99 & 30.70 \\
  MotionGPT~\citep{jiang2024motiongpt} & & \checkmark & & 14.09 & 42.22 & 10.31 & 32.33 \\
  Ours (M2T\&T2M) & & \checkmark & & 15.90 & 42.68 & 11.06 & 33.71 \\
  % \hline
  \midrule
  Ours (TP2T) & \checkmark & & & 11.94 & 41.70 & 9.85 & 31.47 \\
  Ours (V2T) & & & \checkmark & 16.62 & 43.03 & 11.34 & 33.13 \\
  % Ours (MV2T\&T2M) & & \checkmark & \checkmark & \textbf{20.32} & 45.33 & 12.80 & \textbf{35.31} \\
  % \hline
  \midrule
  Ours (TPV2M $+$ MV2T) & \checkmark & & \checkmark & \textbf{19.97} & 45.41 & \textbf{12.81} & \textbf{35.04} \\
  Ours (TPV2T) & \checkmark & & \checkmark & 18.38 & 44.55 & 12.12 & 33.80 \\
  % Ours (TPV2T w/ GPT-2 Large) & \checkmark & & \checkmark & \textbf{19.56} & 44.48 & 12.49 & \textbf{35.21} \\
  Ours (Joint Training) & \checkmark & & \checkmark & 19.40 & \textbf{45.45} & 12.72 & 34.82 \\
  \bottomrule
  \end{tabular}
  \vspace{-5pt}
\end{table}

\begin{figure}[tb]
  \centering
  \includegraphics[width=\textwidth]{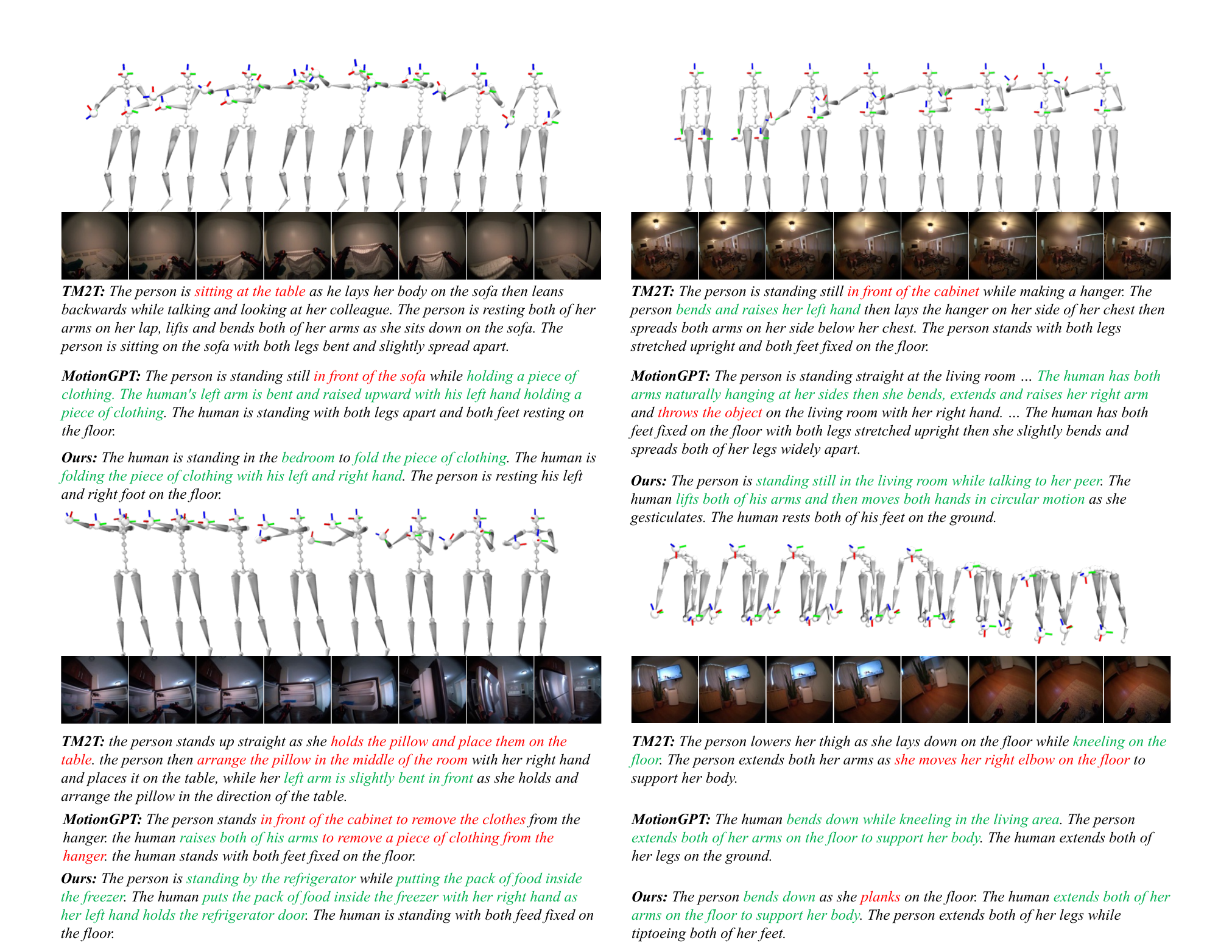}
  \vspace{-15pt}
  \caption{\textbf{Qualitative Results of Motion Understanding.} We use green to highlight correct parts and red for mistakes.}
  \label{fig:motion_understanding}
  \vspace{-15pt}
\end{figure}

% TM2T, MotionGPT, V2T, GT
\noindent\textbf{Qualitative Results.}
We show four examples of motion understanding in Fig.~\ref{fig:motion_understanding}.
% We use green to highlight the correct parts and red to mark the mistakes.
TM2T and MotionGPT use full body motions as inputs. Ours is the full version with three-points and egocentric videos as inputs. TM2T's language generation part is trained from scratch. Therefore, it often makes mistakes about motions and even generates texts that does not make sense. MotionGPT can generate reasonable descriptions for the motions. In the lower left example, just from the motions, ``removing a piece of clothing from the hanger'' is a reasonable answer. However, our target motion narration is highly related to environments. TM2T and MotionGPT fail to generate correct narrations for the lack of vision signals. For our model, even though we do not directly use motions as inputs, \nickname{} jointly model the distributions of different modalities and can generate correct narrations according to different scenarios.

\subsection{Ablation Study}
% window size, vqvae ablation
\noindent\textbf{Window Size of Motion Tracking.}
As shown in Tab.~\ref{tab:win_size}, we increase the window size for three-points motion tracking from $60$ to $120$ frames, which brings an improvement of $4.2mm$ in joint position errors. This is reasonable since increasing the window size brings more contexts, which helps the disambiguation. If we further include egocentric videos in the inputs, the improvement of increasing window size is not as large. Moreover, using $60$ frames plus egocentric video shows better performance than only using $120$ frames. This indicates that the context of egocentric video might be more effective than increasing window size.

\noindent\textbf{Motion VQ-VAE.}
Ablation studies on motion VQ-VAE are reported in Tab.~\ref{tab:vqvae}. ``PQ'' is the number of codebooks. ``CB'' is the total number of entries in codebooks. The first two lines shows that large improvements can be achieved by simply using product quantization. Moreover, increasing the number of codes and decreasing code dimensions bring further improvement.

\noindent\textbf{Larger Language Model.} We use GPT-2 Medium (345M) to conduct most of our experiments for efficiency. To examine the potential of using larger LM, we train with GPT-2 Large (1.5B) and report performance on TPV2T in Tab.~\ref{tab:largerlm}. The improved scores suggest \nickname{}'s scalability as a versatile framework.

% \setlength{\tabcolsep}{2.8pt}
% \begin{table}[h]
%   \vspace{-12pt}
%   \centering
%   \footnotesize
%   \caption{Ablation on the LM Backbone Size.}
%   \label{tab:largerlm}
%   \begin{tabular}{c|cccc}
%   \toprule
%   Model (Size) & Bert$\uparrow$ & Bleu@1$\uparrow$ & Bleu@4$\uparrow$ & RougeL$\uparrow$ \\
%   \midrule
%   GPT-2 Medium (345M) & 18.38 & \textbf{44.55} & 12.12 & 33.80 \\
%   GPT-2 Large (1.5B) & \textbf{19.56} & 44.48 & \textbf{12.49} & \textbf{35.21} \\
%   \bottomrule
%   \end{tabular}
%   \vspace{-12pt}
% \end{table}

% \setlength{\tabcolsep}{2.8pt}
% \begin{table}[h]
%   \vspace{-12pt}
%   \centering
%   \footnotesize
%   \caption{Ablation on the LM Backbone Size.}
%   \label{tab:largerlm}
%   \begin{tabular}{c|cccc}
%   \toprule
%   Model (Size) & Bert$\uparrow$ & Bleu@1$\uparrow$ & Bleu@4$\uparrow$ & RougeL$\uparrow$ \\
%   \midrule
%   GPT-2 Medium (345M) & 18.38 & \textbf{44.55} & 12.12 & 33.80 \\
%   GPT-2 Large (1.5B) & \textbf{19.56} & 44.48 & \textbf{12.49} & \textbf{35.21} \\
%   \bottomrule
%   \end{tabular}
%   \vspace{-12pt}
% \end{table}

\setlength{\tabcolsep}{2.8pt}
\begin{table}[t]
\vspace{-15pt}
\parbox{.3\linewidth}{
\centering
\scriptsize
\caption{Ablation Study on Window Size for Motion Tracking.}
\label{tab:win_size}
\vspace{-5pt}
\begin{tabular}{cc|cccc}
    \toprule
    Win & Vid & Full & Upper & Lower & J.A. \\
    \midrule
    60 & & 83.88 & 54.06 & 148.37 & 13.31 \\
    120 & & 79.61 & 52.66 & 138.87 & 13.01 \\
    60 & \checkmark & 73.38 & 49.67 & 124.58 & \textbf{12.48} \\
    120 & \checkmark & \textbf{72.76} & \textbf{49.20} & \textbf{123.09} & 12.52 \\
    \bottomrule
\end{tabular}
}
\hfill
\parbox{.35\linewidth}{
\centering
\scriptsize
\caption{Ablation Study on Reconstruction Results of Motion VQ-VAE. [$mm$]}
\vspace{-5pt}
\label{tab:vqvae}
\begin{tabular}{ccc|ccc}
    \toprule
    PQ & CB & Dim & \scriptsize{MPJPE} & \scriptsize{PA-MPJPE} & \scriptsize{ACCEL} \\
    \midrule
    1 & 2048 & 512 & 51.60 & 37.55 & 1.09 \\
    2 & 2048 & 512 & 39.63 & 29.77 & 0.71 \\
    % 2 & 4096 & 512 & 39.20 & 29.66 & 0.82 \\
    2 & 16384 & 256 & 39.13 & 29.78 & 1.08 \\
    2 & 16384 & 64 & \textbf{34.49} & \textbf{26.83} & \textbf{0.67} \\
    \bottomrule
\end{tabular}
}
\hfill
\parbox{.25\linewidth}{
\centering
\scriptsize
\caption{Ablation on the LM size. Medium: 345M; Large: 1.5B}
  \label{tab:largerlm}
  \vspace{-5pt}
  \begin{tabular}{c|cc}
  \toprule
  GPT-2 Size & Medium & Large \\
  \midrule
  Bert$\uparrow$ & 18.38 & \textbf{19.56}\\
  Bleu@1$\uparrow$ & \textbf{44.55} & 44.48\\
  Bleu@4$\uparrow$ & 12.12 & \textbf{12.49}\\
  RougeL$\uparrow$ & 33.80 & \textbf{35.21}\\
  \bottomrule
  \end{tabular}
}
\vspace{-5pt}
\end{table}

\subsection{More Applications}

\begin{figure}[tb]
  \centering
  \includegraphics[width=\textwidth]{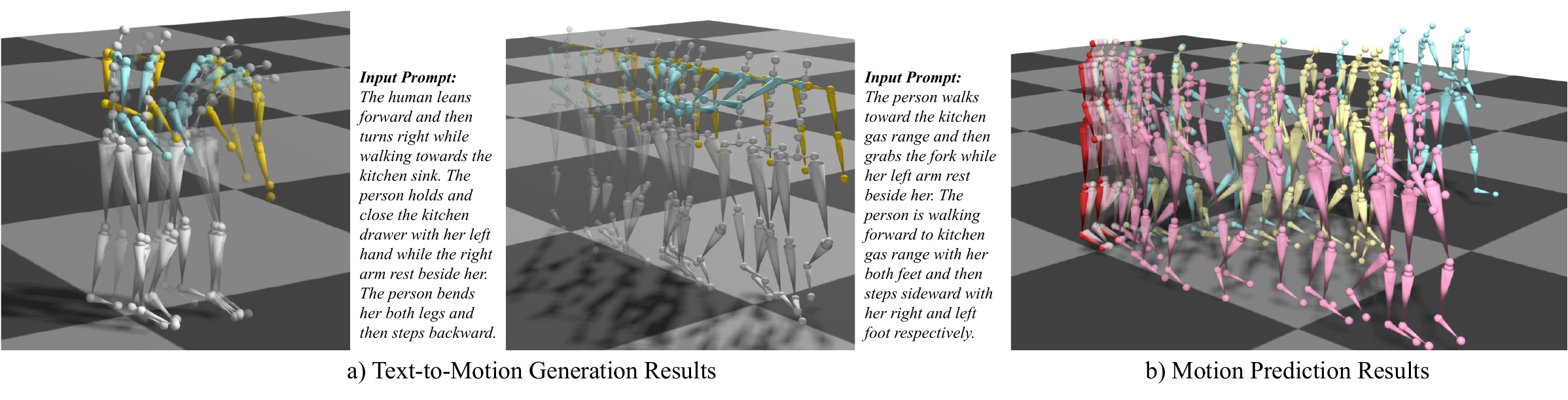}
  \vspace{-20pt}
  \caption{\textbf{More Analysis on \nickname{}.} \textbf{a)} Qualitative results of text-to-motion generation. \textbf{b)} Qualitative results of motion prediction.}
  \label{fig:analysis}
  \vspace{-15pt}
\end{figure}

\noindent\textbf{Text-to-Motion Generation.} As part of our joint training, \nickname{} is capable of generating motions from texts, as shown in Fig.~\ref{fig:analysis} a). Even with long prompts separately describing upper body and lower body, our model is able to generate motions that match the inputs.

\noindent\textbf{Motion Prediction.} As a by-product of the motion pre-training, \nickname{} can function as a motion predictor. As shown in Fig.~\ref{fig:analysis} b), given motion prompts (the red skeleton in the left), subsequent motions can be randomly sampled. We show three different samples in different colors.

\section{Discussion}
\label{sec:discussion}
We propose \nickname{}, a multi-modal language model for egocentric motion tracking and understanding.
% A motion VQ-VAE is trained as the motion tokenizer. Motion pre-training is conducted for the language model to learn the distribution of motions. The multi-modal instruction tuning is performed for downstream tasks.
A three-steps paradigm, including motion tokenization, motion pre-training and multi-modal instruction tuning, is proposed to facilitate the training.
In contrast to previous works, the proposed framework unifies the egocentric motion tasks with a language model, and incorporates multi-modal sensor data as context information, which is proven effective for both tasks.

\noindent\textbf{Limitations.} Firstly, our motion tokenizer is a VQ-VAE, which carries reconstruction errors. It sets an upper bound for motion tracking.
Moreover, for the motion tracking training, the loss is calculated on discrete motion tokens, instead of raw motion representations, which might also harm the performance of motion tracking.
Secondly, for motion understanding, since each egocentric video frame is compressed by the CLIP encoder to a one-dimensional vector, it is hard for models to precisely name the object that the person is interacting with. Moreover, as is commonly observed in language models~\citep{ji2023survey}, \nickname{} also suffers from the hallucination problem.

\noindent\textbf{Potential Societal Impact.} While contextual AI offers opportunities for efficiency improvement and societal advancement, the collection and analysis of human data could lead to privacy issues for both users and people around.

% \clearpage  % TODO REVIEW/FINAL: This \clearpage needs to be removed from both review and camera-ready versions.
\clearpage


\bibliographystyle{iclr2024_conference}
\bibliography{egbib}

\clearpage
\renewcommand{\thesection}{\Alph{section}} 
\setcounter{section}{0}
\section*{\Large Supplementary}\label{supp}
We provide more implementation details (Sec.~\ref{sec:supp_exp}) and qualitative results (Sec.~\ref{sec:supp_res}) in this supplementary material. To better showcase our results, We also provide videos  in our project page \url{https://hongfz16.github.io/projects/EgoLM}.

\begin{figure}[h]
  \centering
  \includegraphics[width=\textwidth]{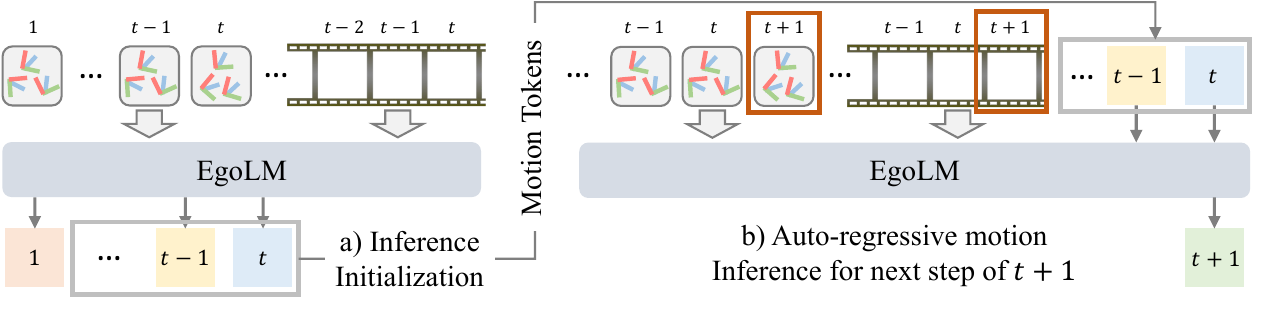}
  \vspace{-15pt}
  \caption{\textbf{Online Motion Tracking Inference.} For the new time step of $t+1$ with new data coming in, last motion tokens are combined with the new input tokens to decode the next motion token $t+1$.}
  \label{fig:ae_motion_tracking}
  \vspace{-15pt}
\end{figure}

\section{Implementation Details} \label{sec:supp_exp}

\subsection{Auto-regressive Inference for Motion Tracking}

At inference time, motion understanding is the same as the language model inference. For motion tracking, it usually requires online inference over a long period. With a language model, which is an auto-regressive model, it is straight-forward to perform online motion tracking. As shown in Fig.~\ref{fig:ae_motion_tracking}, firstly, an initialization over the first $t$ frames of data is required. When the new data frame $t+1$ comes in, the input conditions are updated accordingly. Then, it is not necessary to predict all the motion tokens from frame $2$ to frame $t+1$. We take the previously generated motion tokens from frame $2$ to frame $t$ as inputs and prompt the network to generate one more token for frame $t+1$.

\subsection{Evaluation Metrics}

For motion tracking, we use joint position errors and joint angle errors to evaluate the performance. Specifically, for the joint position errors, we first align ground truth skeletons and generated skeletons by the head positions only by translation. Then full body, upper body and lower body joint position errors are calculated separately. Joint angle errors are calculated on full body and root joints.
For the evaluation of motion VQ-VAE in main paper Tab. 4, we apply widely adopted metrics for motion regression, \ie, Mean Per-Joint Position Error (MPJPE)~\citep{ionescu2013human3}, Procrustes-Aligned (PA-)MPJPE~\citep{hmrKanazawa17}, and joint position acceleration (ACCL) error.
For the motion understanding, we use standard NLP metrics, please kindly refer to corresponding papers for more details.
% \lingni{Shall we also mention the motion understandign metrics} NLP metrics are pretty standard and already explained in the main paper.

% \subsection{Dataset Details}
% We use a self-collected large-scale multi-modal motion dataset for the experiments of this work. The detailed report of the dataset, \texttt{dataset\_report.pdf}, is included as part of our supplementary material. Please kindly refer to the report for more details of the dataset. In this paper, we use a subset of the above-mentioned dataset, which were collected before November 2023. Three-points and one-point inputs are mocked up from the ground truth motions by taking the 6-DoF poses of head and both hands. The narrations are collected by asking human annotators to give descriptions for full body, upper body and lower body motions. Therefore, our target motion narrations consists of all three aspects of the motion.

\section{More Qualitative Results} \label{sec:supp_res}

\subsection{Three-Points Motion Tracking}

\begin{figure}[tb]
  \centering
  \includegraphics[width=\textwidth]{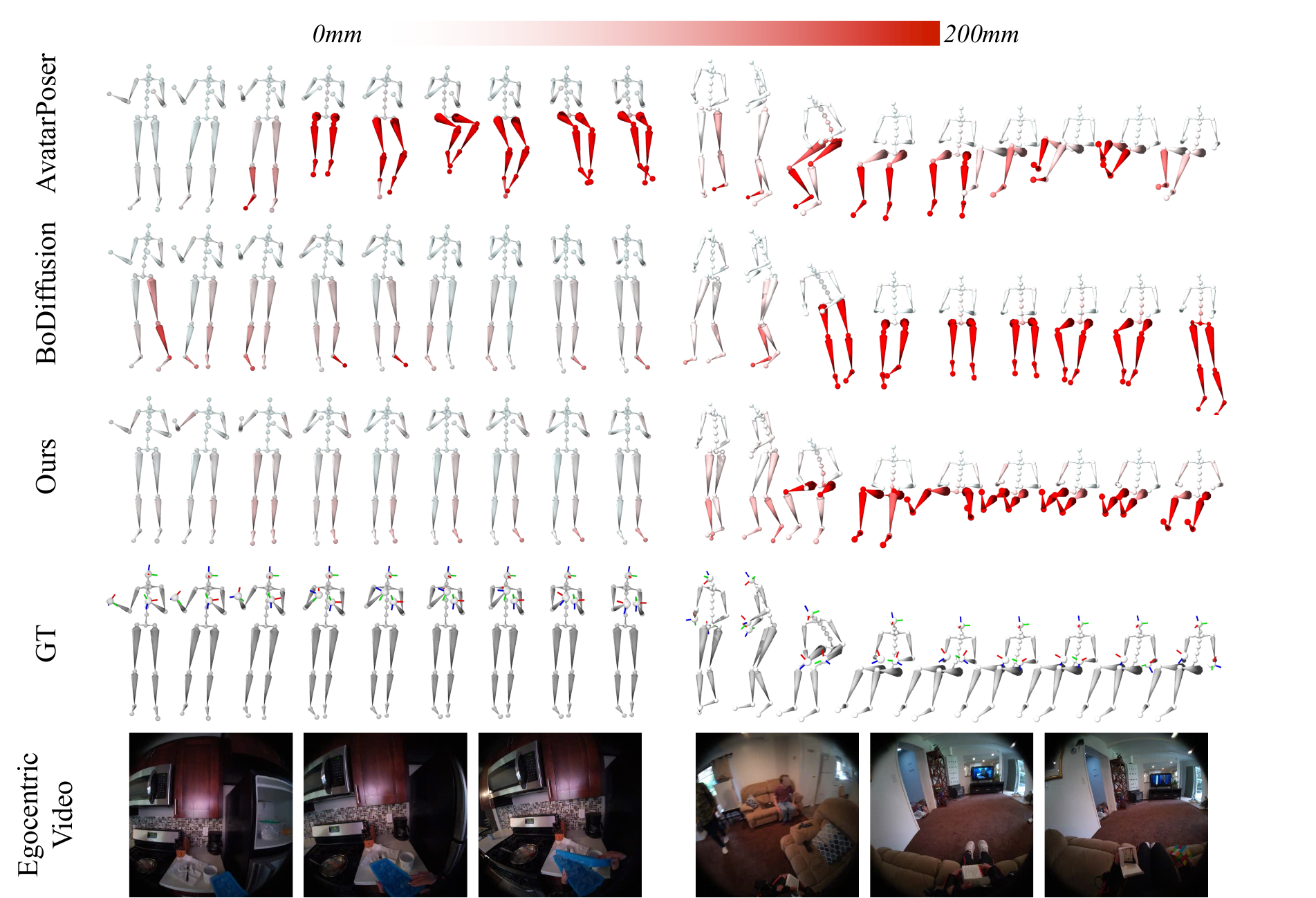}
  \caption{\textbf{Qualitative Results of Three-Points Motion Tracking.} Skeletons are color-coded by joint position errors.}
  \label{fig:threepoints_motion_tracking_0}
\end{figure}

\begin{figure}[tb]
  \centering
  \includegraphics[width=\textwidth]{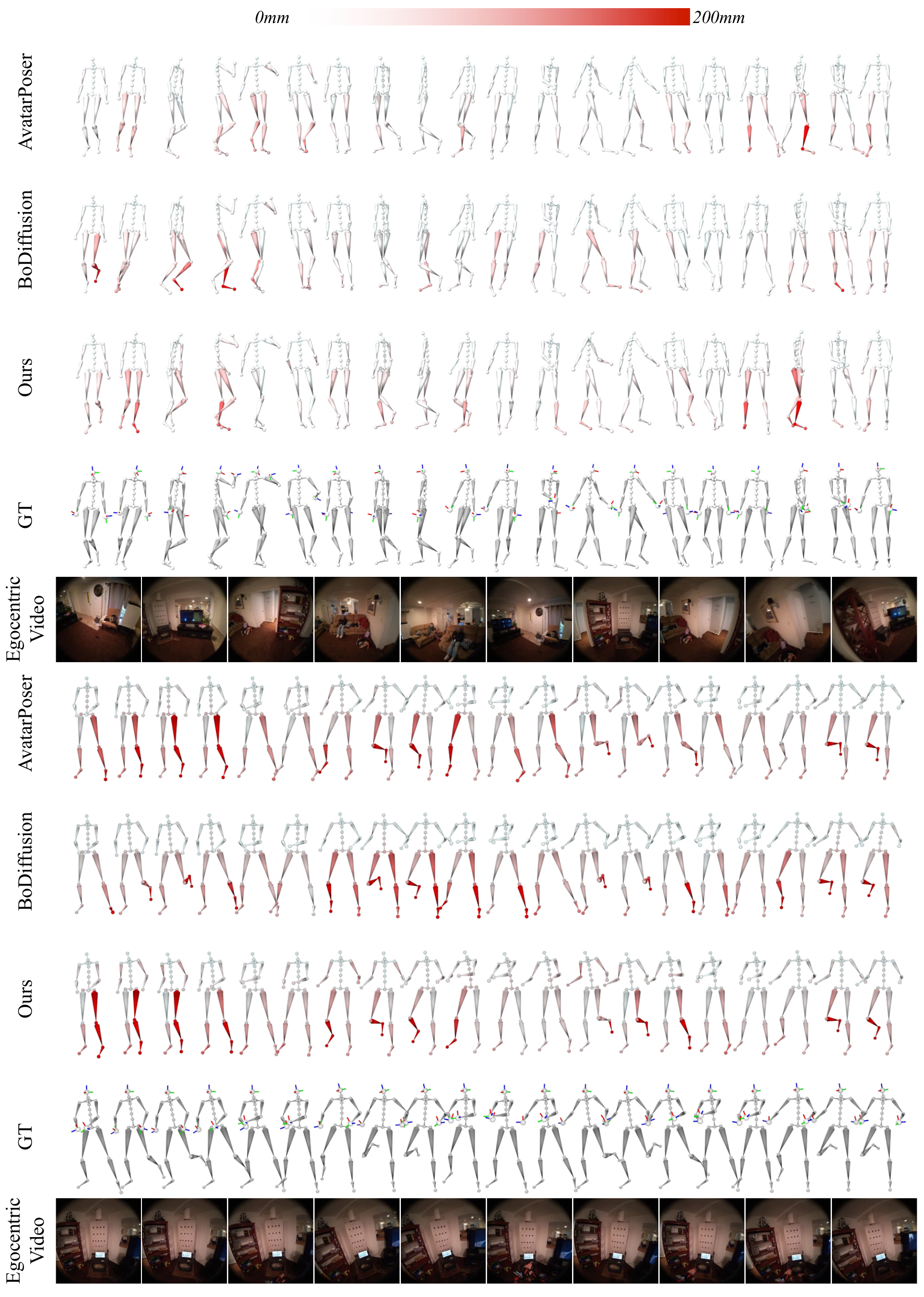}
  \caption{\textbf{Qualitative Results of Three-Points Motion Tracking.} Skeletons are color-coded by joint position errors.}
  \label{fig:threepoints_motion_tracking_1}
\end{figure}

\begin{figure}[tb]
  \centering
  \includegraphics[width=\textwidth]{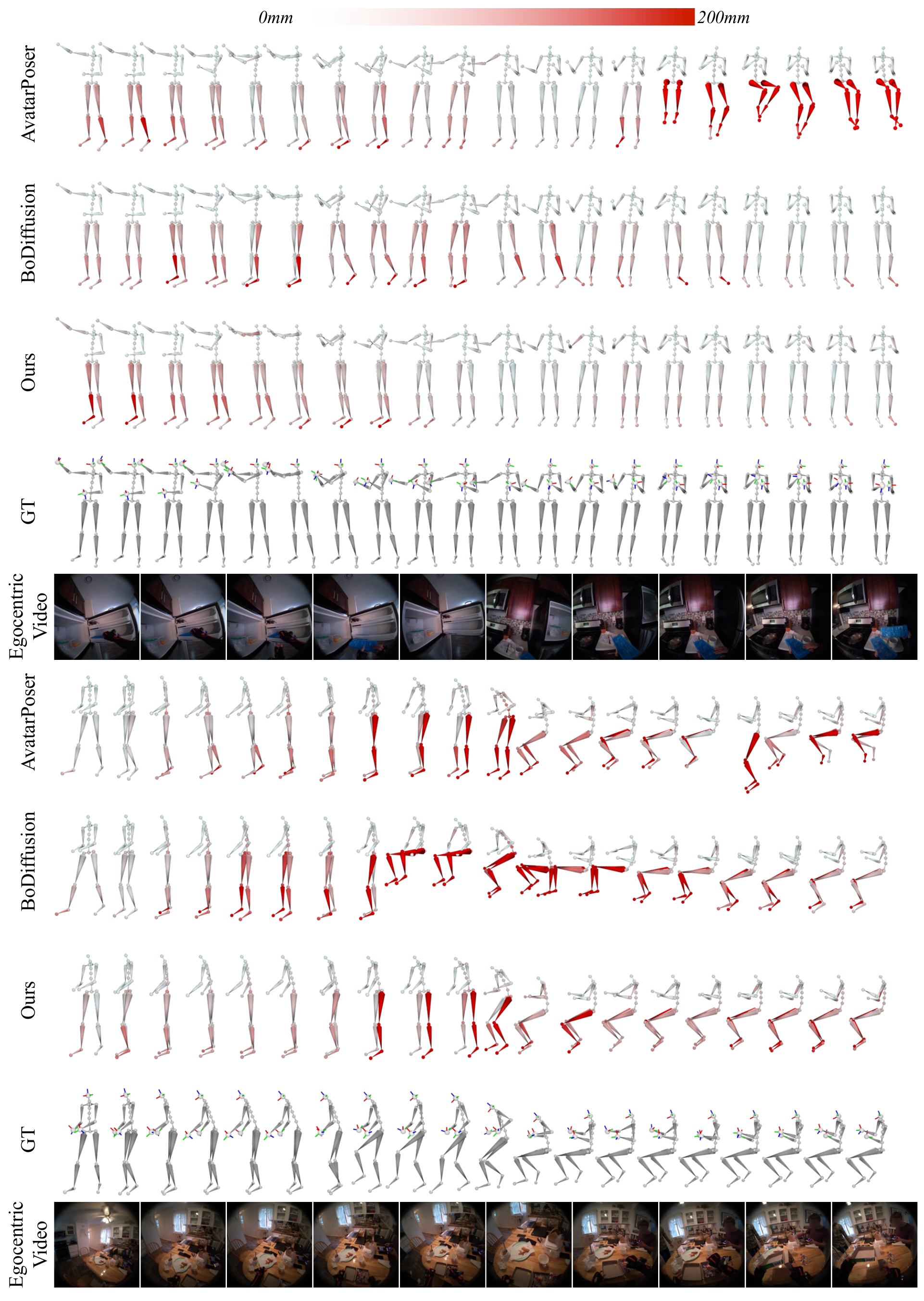}
  \caption{\textbf{Qualitative Results of Three-Points Motion Tracking.} Skeletons are color-coded by joint position errors.}
  \label{fig:threepoints_motion_tracking_2}
\end{figure}

We show four more visual examples of three-points motion tracking in Fig.~\ref{fig:threepoints_motion_tracking_0}, Fig.~\ref{fig:threepoints_motion_tracking_1} and Fig.~\ref{fig:threepoints_motion_tracking_2}. AvatarPoser~\citep{avatarposer} and BoDiffusion~\citep{bodiffusion} are solid baselines that perform well on easy walking cases, \eg, upper example in Fig.~\ref{fig:threepoints_motion_tracking_1}. For the workout sequence, \ie, lower example in Fig.~\ref{fig:threepoints_motion_tracking_2}, even only given three points of upper body, the distribution of lower body motion can be collapsed and generate reasonable motions that matches the ground truth. In Fig.~\ref{fig:threepoints_motion_tracking_2}, we demonstrate the effectiveness of including egocentric videos as inputs. Without any environment context, AvatarPoser and BoDiffusion often fail to distinguish standing and sitting down. We do not assume the knowledge of the head height over the floor, meaning that the three-points positions are normalized to the local coordinates of the first frame. Therefore, it is hard for baseline methods to disambiguate certain scenarios. We propose to introduce contexts using egocentric videos, which contains rich information about the environment and how the person is interacting with it. Therefore, our model can generate the most accurate motions by utilizing these information. For more visualization of three-points motion tracking, please kindly refer to our supplementary videos.
% \lingni{do you mean supp video here? }

\clearpage

\subsection{One-Point Motion Tracking}

\begin{figure}[tb]
  \centering
  \includegraphics[width=\textwidth]{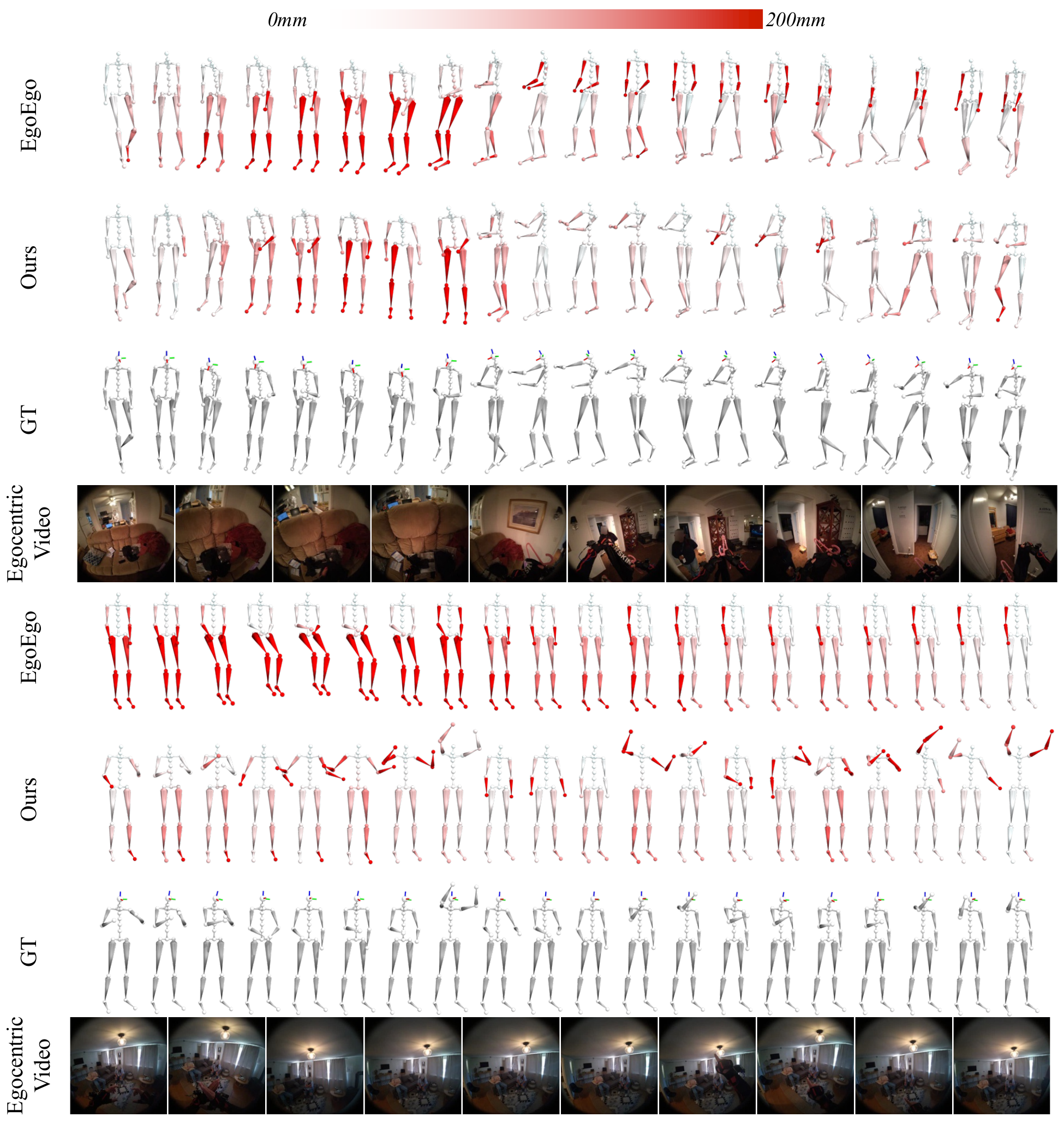}
  \caption{\textbf{Qualitative Results of One-Point Motion Tracking.} Skeletons are color-coded by joint position errors.}
  \label{fig:onepoint_motion_tracking_1}
\end{figure}

\begin{figure}[tb]
  \centering
  \includegraphics[width=\textwidth]{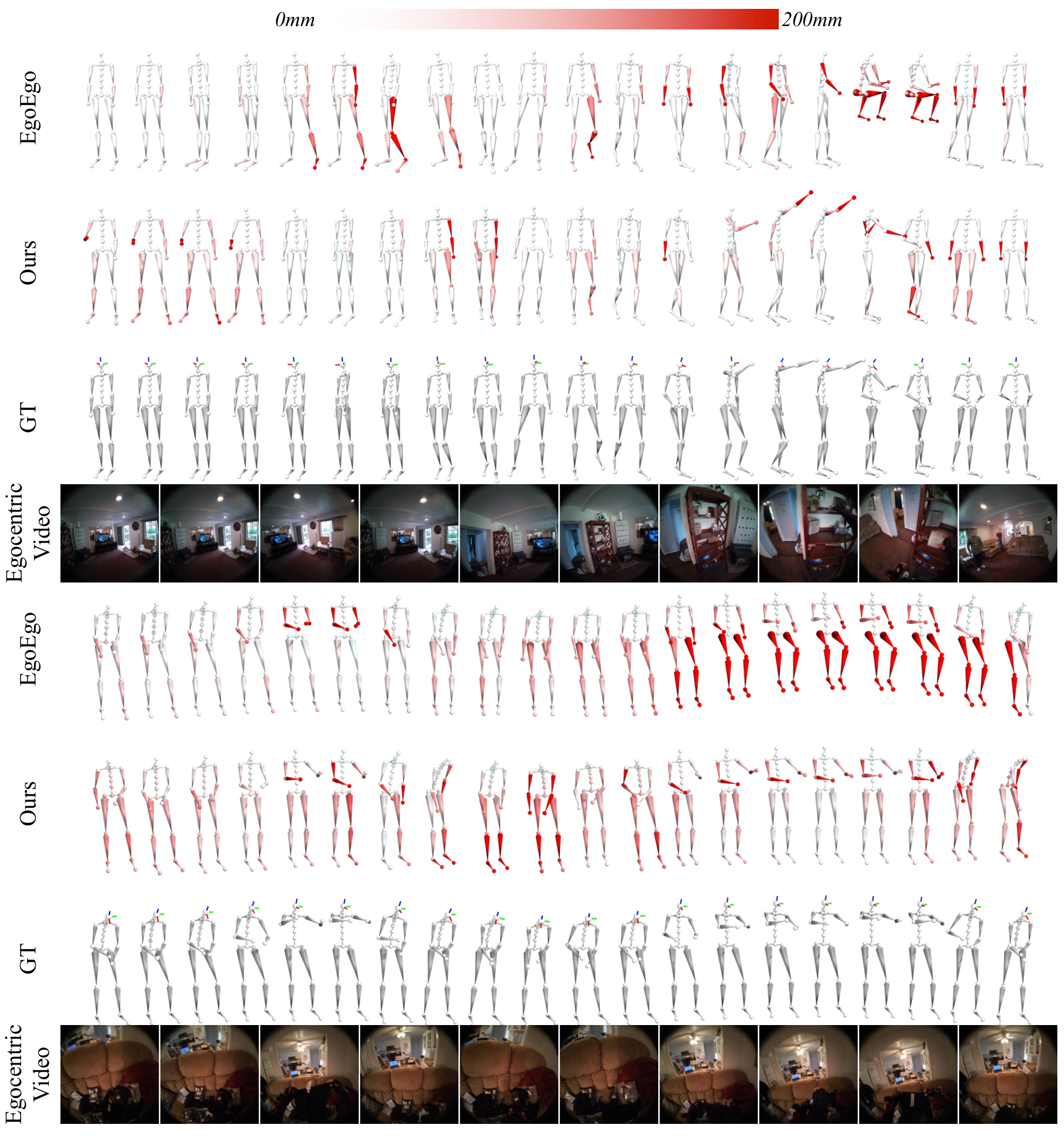}
  \caption{\textbf{Qualitative Results of One-Point Motion Tracking.} Skeletons are color-coded by joint position errors.}
  \label{fig:onepoint_motion_tracking_2}
\end{figure}

We show four more examples of one-point motion tracking in Fig.~\ref{fig:onepoint_motion_tracking_1} and Fig.~\ref{fig:onepoint_motion_tracking_2}. The introduction of egocentric videos has two advantages. Firstly, similar to the case in three-points body tracking, the environment contexts in egocentric videos can disambiguate cases like standing and sitting. Secondly, specifically for one-point motion tracking, egocentric videos provide clues of hand positions. As shown in all four examples, when the person raises the arms in front of the body, hands would be visible in the egocentric videos, which helps the hand position tracking. Admittedly, high-level semantic information provided by CLIP~\citep{radford2021learning} encoders cannot accurately track hand positions. Therefore, as shown in the lower example in Fig.~\ref{fig:onepoint_motion_tracking_1}, our method correctly generates arms moving in the air, but lacks accuracy. For more visual examples of one-point motion tracking, please kindly refer to our supplementary video.

\clearpage

\subsubsection{Multiple Samples.}

\begin{figure}[tb]
  \centering
  \includegraphics[width=\textwidth]{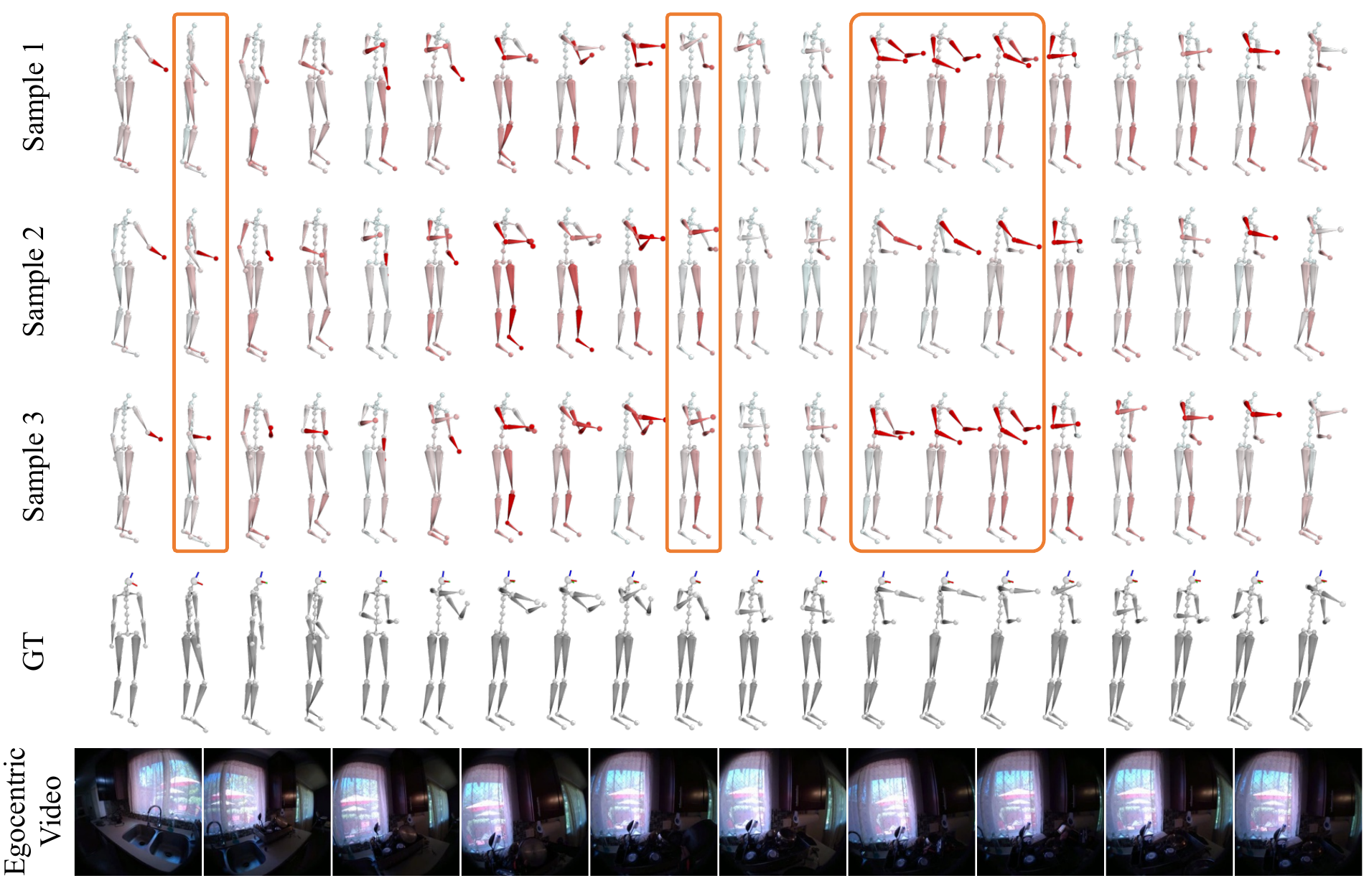}
  \caption{\textbf{Three Random Samples of One-Point Motion Tracking with Egocentric Videos as Inputs.} Since we use language models as our backbone, EgoLM has the ability to randomly sample outputs given the same inputs. Egocentric videos provide strong clues for hand positions, leading to less diversity as shown in the highlighted areas.}
  \label{fig:onepointclip_motion_tracking_multisample}
\end{figure}

Note that EgoLM is essentially a generative model. Therefore, our model is capable of generating different samples with the same inputs. In Fig.~\ref{fig:onepointclip_motion_tracking_multisample}, we show three random samplings on the same input one-point and egocentric video. When hands are not visible in the frame, \ie, the left highlighted frame, hand positions are not constrained, and therefore shows high diversity across different samples. For the other highlighted frames, hands are visible in the egocentric videos, which helps to collapse the distribution of possible positions of hands. But as discussed above, our way of encoding egocentric videos cannot accurately track the hand positions. Therefore, our model also shows some diversity of hand positions in these cases.

\begin{figure}[tb]
  \centering
  \includegraphics[width=\textwidth]{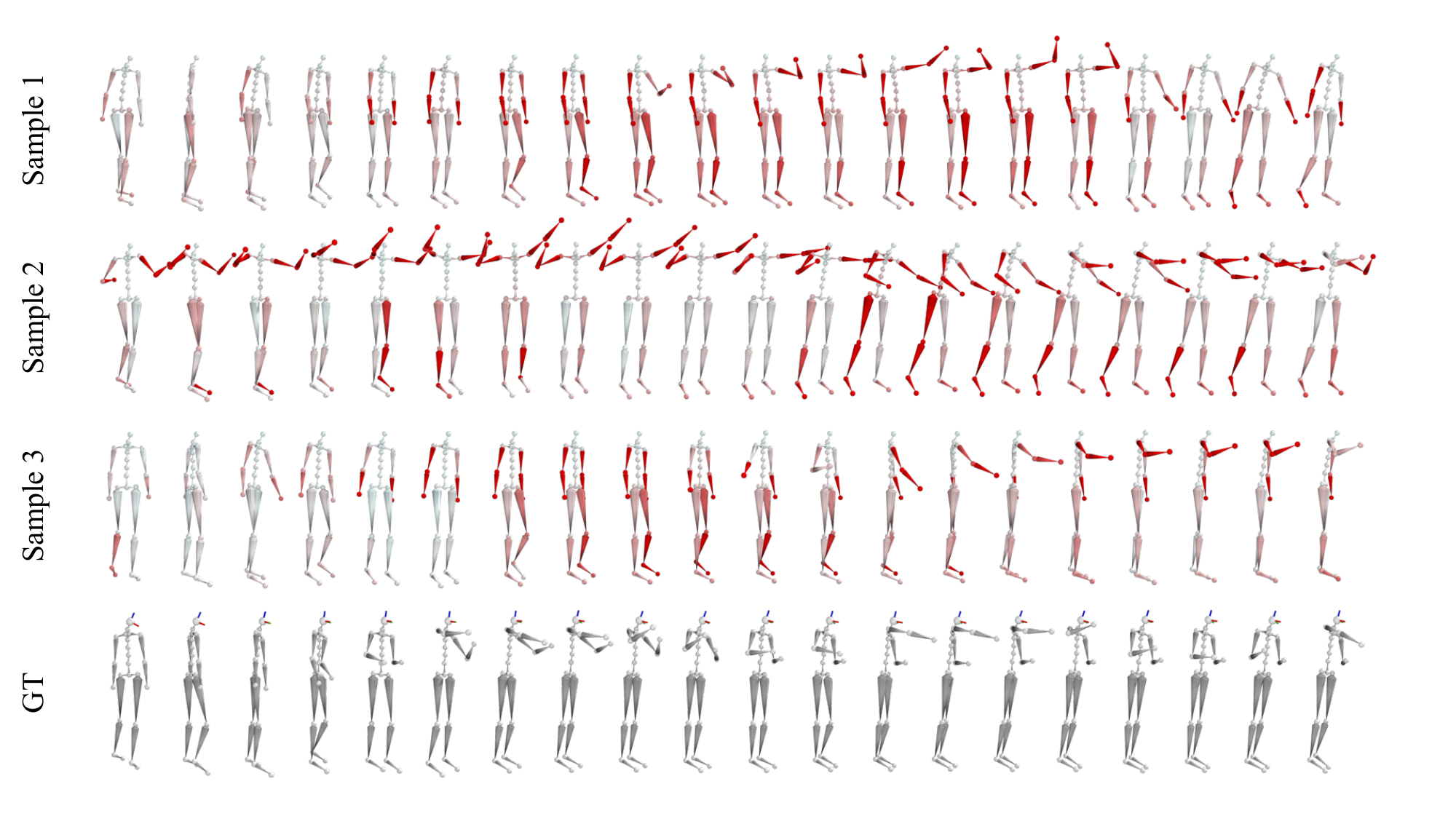}
  \caption{\textbf{Three Random Samples of One-Point Motion Tracking without Egocentric Videos as Inputs.} With only head poses as inputs, the generation of full body motion, especially upper body motions, is less constrained.}
  \label{fig:onepoint_motion_tracking_multisample}
\end{figure}

To further demonstrate the diversity of our model, we also show three random samples from our one-point motion tracking model that does not take egocentric videos as inputs in Fig.~\ref{fig:onepoint_motion_tracking_multisample}. Lack of any indication of the hand positions, the upper body generation is even less constrained than that of the lower body and shows high diversity across three samples.

\clearpage

\subsection{Motion Understanding}

\begin{figure}[tb]
  \centering
  \includegraphics[width=\textwidth]{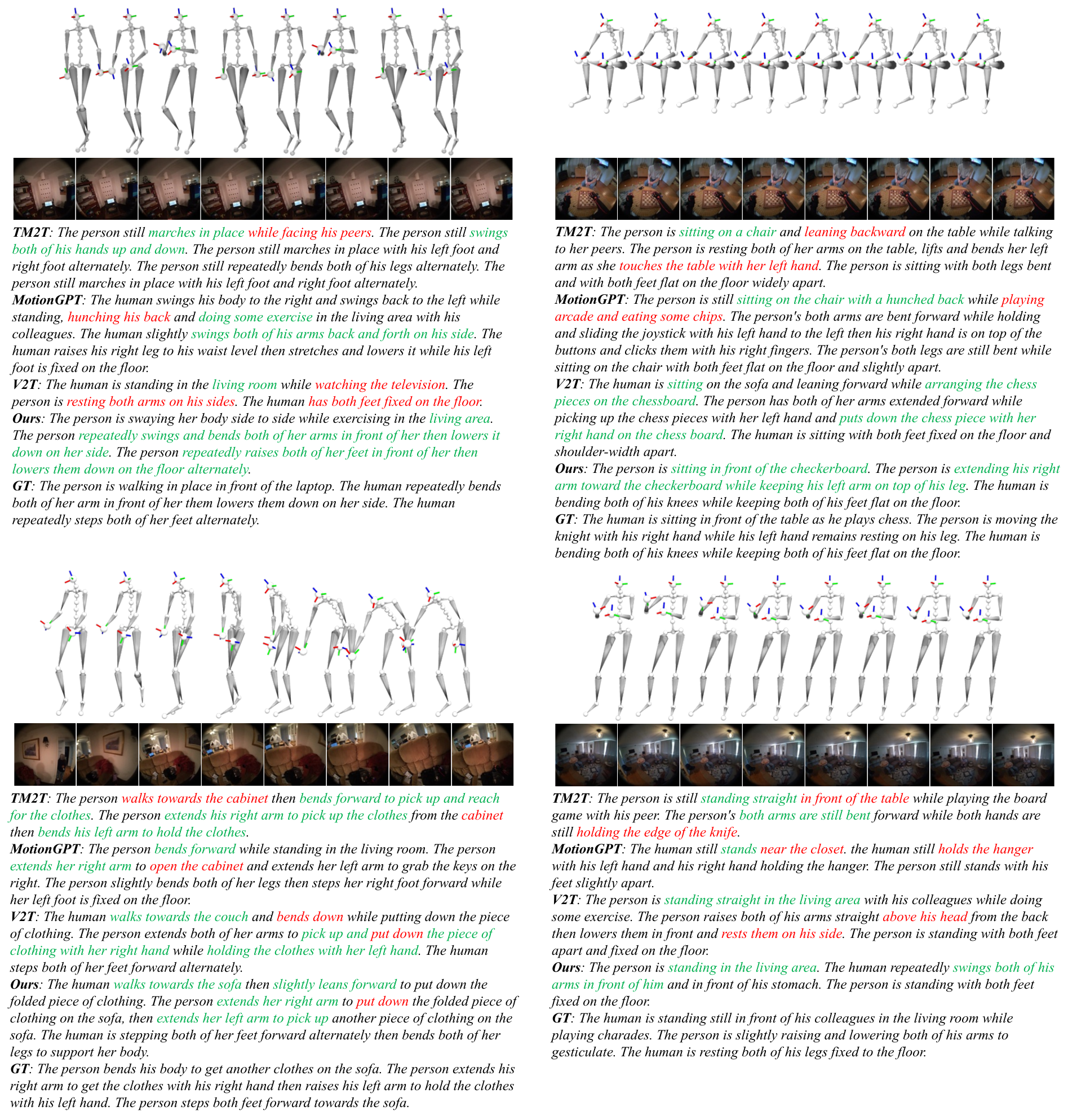}
  \caption{\textbf{Qualitative Results of Motion Understanding.} We use green to highlight correct parts in the answers while red for wrong ones.}
  \label{fig:motion_understanding_1}
\end{figure}

\begin{figure}[tb]
  \centering
  \includegraphics[width=\textwidth]{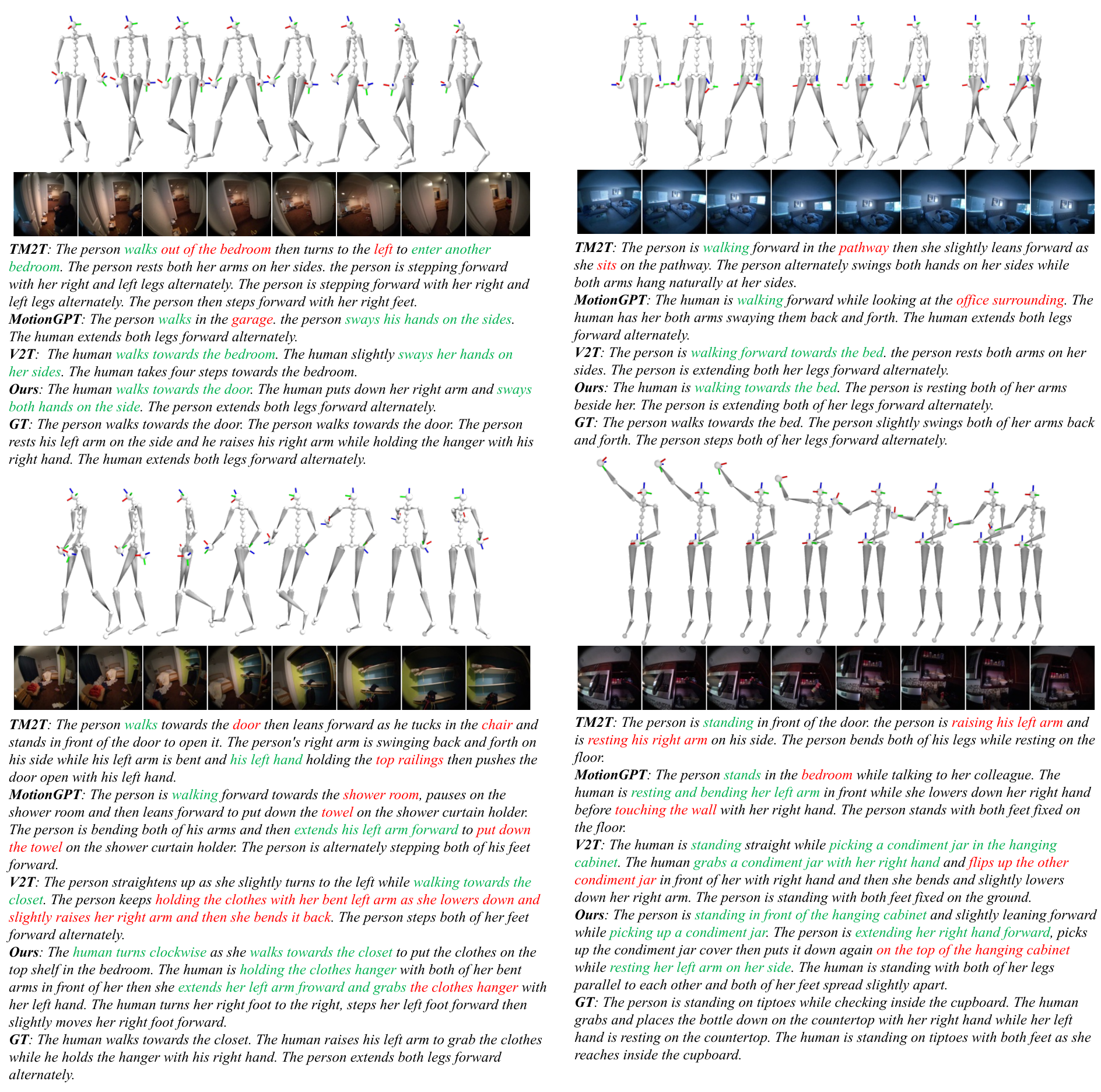}
  \caption{\textbf{Qualitative Results of Motion Understanding.} We use green to highlight correct parts in the answers while red for wrong ones.}
  \label{fig:motion_understanding_2}
\end{figure}

We show eight more examples of motion understanding in Fig.~\ref{fig:motion_understanding_1} and Fig.~\ref{fig:motion_understanding_2}. Similar to the main paper, we use green to highlight correct parts in the answers and red for wrong answers. Similar to the observation made in the main paper, even though TM2T~\citep{guo2022tm2t} and MotionGPT~\citep{jiang2024motiongpt} have access to the full body motion, the generated narrations are reasonable but completely wrong if consider the environment context. For example, in the upper right example in Fig.~\ref{fig:motion_understanding_2}, given the simple walking sequence, both TM2T and MotionGPT can correctly understanding that the person is walking forward. But they all give the wrong answers about the places the person is walking in. Thanks to the egocentric videos, our model successfully produces the correct description as ``walking towards the beds''.

\clearpage

\subsection{Motion Prediction}

\begin{figure}[tb]
  \centering
  \includegraphics[width=\textwidth]{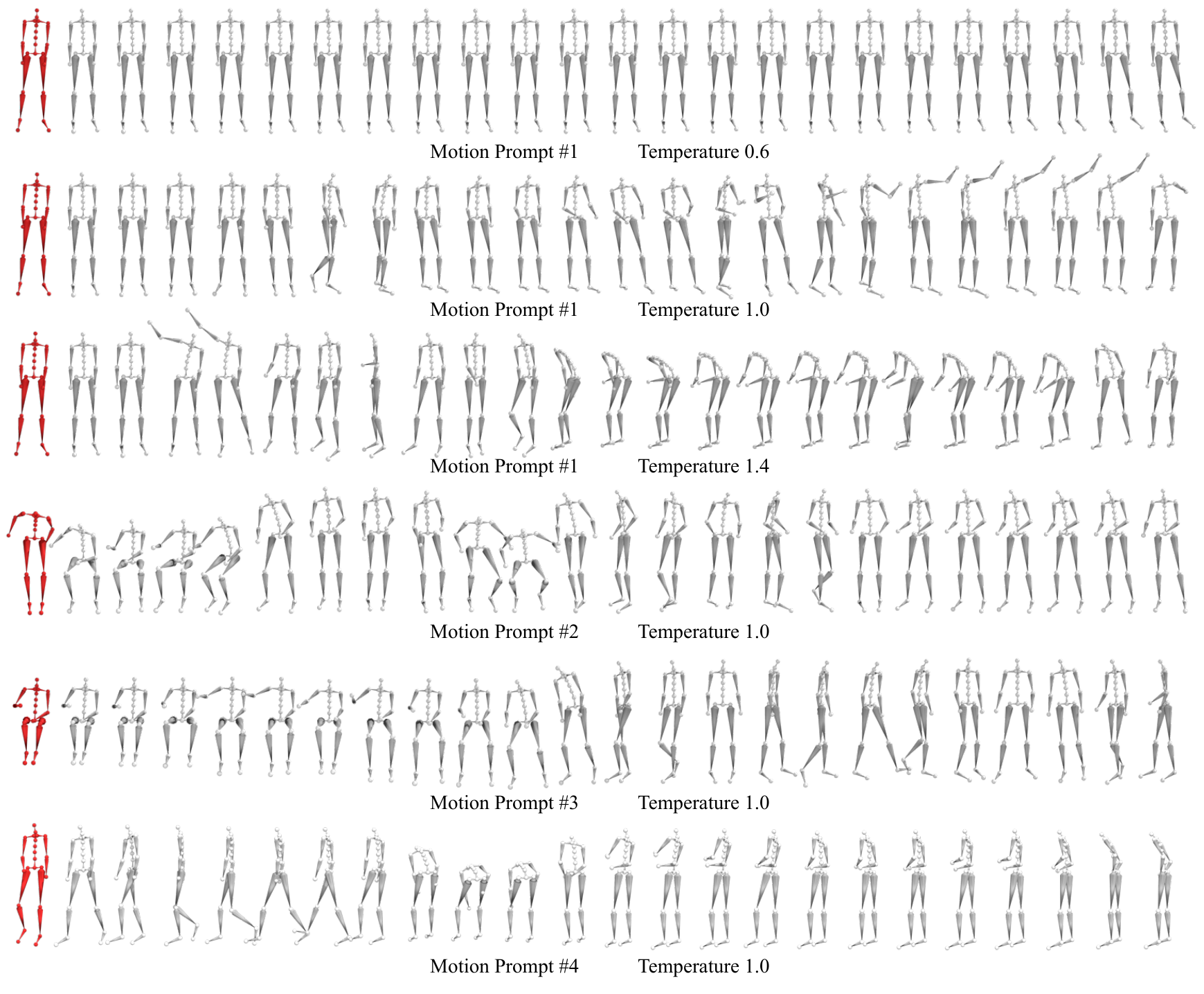}
  \caption{\textbf{Qualitative Results of Motion Prediction.} The first skeletons in red are input motion prompts. The following motions are randomly sampled auto-regressively from our motion pre-training network.}
  \label{fig:motion_prediction}
\end{figure}

As a by-product of the second stage of our training pipeline, motion pre-training, we build a motion prediction network. Given leading motions as the prompts, our model is capable of auto-regressively sample motions that complete the motion prompts. As shown in Fig.~\ref{fig:motion_prediction}, the first three samples show three different samples given the same motion prompt. We can increase the intensity of the generated motions by increasing the temperature. The last three samples show three random samples given various motion prompts, \eg, bending forward, sitting down and standing.

\end{document}